%% file: main.tex
\documentclass[10pt,twocolumn,letterpaper]{article}

\usepackage[pagenumbers]{wacv} 

\usepackage{graphicx}
\usepackage{amsmath}
\usepackage{amssymb}
\usepackage{booktabs}
\usepackage[nolist,nohyperlinks]{acronym}
\usepackage{bm}
\usepackage{bbm}
\usepackage{amsfonts}
\usepackage{svg} 
\usepackage[nolist,nohyperlinks]{acronym}
\usepackage{enumitem}
\usepackage{caption}
\usepackage{subcaption} 
\usepackage{url}
\usepackage{pifont}
\newcommand{\cmark}{\ding{51}}  
\newcommand{\xmark}{\ding{55}}  
\usepackage[accsupp]{axessibility}  

\graphicspath{{images/}}

\usepackage{array,booktabs}
\newcolumntype{M}[1]{>{\centering\arraybackslash}m{#1}} 
\usepackage{arydshln}

\makeatletter
\def\adl@drawiv#1#2#3{%
        \hskip.5\tabcolsep
        \xleaders#3{#2.5\@tempdimb #1{1}#2.5\@tempdimb}%
                #2\z@ plus1fil minus1fil\relax
        \hskip.5\tabcolsep}
\newcommand{\cdashlinelr}[1]{%
  \noalign{\vskip\aboverulesep
           \global\let\@dashdrawstore\adl@draw
           \global\let\adl@draw\adl@drawiv}
  \cdashline{#1}
  \noalign{\global\let\adl@draw\@dashdrawstore
           \vskip\belowrulesep}}
\makeatother

\usepackage{hyperref}
\hypersetup{breaklinks,colorlinks}

\usepackage[capitalize]{cleveref}
\crefname{section}{Sec.}{Secs.}
\Crefname{section}{Section}{Sections}
\Crefname{table}{Table}{Tables}
\crefname{table}{Tab.}{Tabs.}

\def\confName{WACV}
\def\confYear{2024}

\newcommand{\ourmodelname}{ContrasTR} 
\newcommand{\paragrax}[1]{{\bf #1}\quad}

\begin{document}

\begin{acronym}[ICANN]
    \acro{mot}[MOT]{Multi-Object Tracking} 
    \acro{cnn}[CNN]{Convolutional Neural Network}
    \acro{mota}[MOTA]{Multi-Object Tracking Accuracy}
    \acro{mmota}[mMOTA]{mean Multi-Object Tracking Accuracy}
    \acro{motp}[MOTP]{Multi Object Tracking Precision}
    \acro{hota}[HOTA]{Higher Order Tracking Accuracy}
    \acro{mhota}[mHOTA]{mean Higher Order Tracking Accuracy}
    \acro{idf1}[IDF1]{Identity F1 Score}
    \acro{midf1}[mIDF1]{mean Identity F1 Score}
    \acro{ffn}[FFN]{Feed-Forward Network}
    \acrodefplural{cnn}[CNNs]{Convolutional Neural Networks} 
\end{acronym}

\title{Contrastive Learning for Multi-Object Tracking with Transformers}

\author{%
    Pierre-François De Plaen*$^{1}$ \and Nicola Marinello*$^{1}$ \and Marc Proesmans$^{1,3}$ \and Tinne Tuytelaars$^{1}$ \and Luc Van Gool$^{1,2,3}$\\
    $^1$ ESAT-PSI, KU Leuven, Belgium \quad $^2$ CVL, ETH Zürich, Switzerland \quad $^3$ TRACE vzw \\
    \small{\texttt{\{pdeplaen,nicola.marinello,marc.proesmans,tinne.tuytelaars,luc.vangool\}@esat.kuleuven.be}}
}
\maketitle

\def\thefootnote{*}
\begin{NoHyper}
\footnotetext{These authors contributed equally.}
\end{NoHyper}
\def\thefootnote{\arabic{footnote}}

\begin{abstract}
    The DEtection TRansformer (DETR) opened new possibilities for object detection by modeling it as a translation task: converting image features into object-level representations. Previous works typically add expensive modules to DETR to perform \ac{mot}, resulting in more complicated architectures. 
    We instead show how DETR can be turned into a \ac{mot} model by employing an instance-level contrastive loss, a revised sampling strategy and a lightweight assignment method. Our training scheme learns object appearances while preserving detection capabilities and with little overhead. Its performance surpasses the previous state-of-the-art by +2.6 mMOTA on the challenging BDD100K dataset and is comparable to existing transformer-based methods on the MOT17 dataset.
\end{abstract}

\input{intro.tex}
\input{related_work.tex}

\input{model.tex}
\input{experiments.tex}
\input{conclusion.tex}

{\small
\bibliographystyle{ieee_fullname}
\bibliography{egbib}
}


\clearpage
\appendix
\input{supplementary.tex}

\end{document}

%% file: intro.tex
\section{Introduction}

In recent years, transformer networks \cite{vaswani2017attention,dosovitskiy2020image,liu2021swin} have gained momentum in computer vision.
The original transformer \cite{vaswani2017attention} was designed for sequence modeling and transduction tasks. Its architecture is notable for using attention modules, which allow for capturing long-range contextual relationships between word tokens.

\begin{figure}[htp]
\centering
        \includegraphics[width=0.47\textwidth]{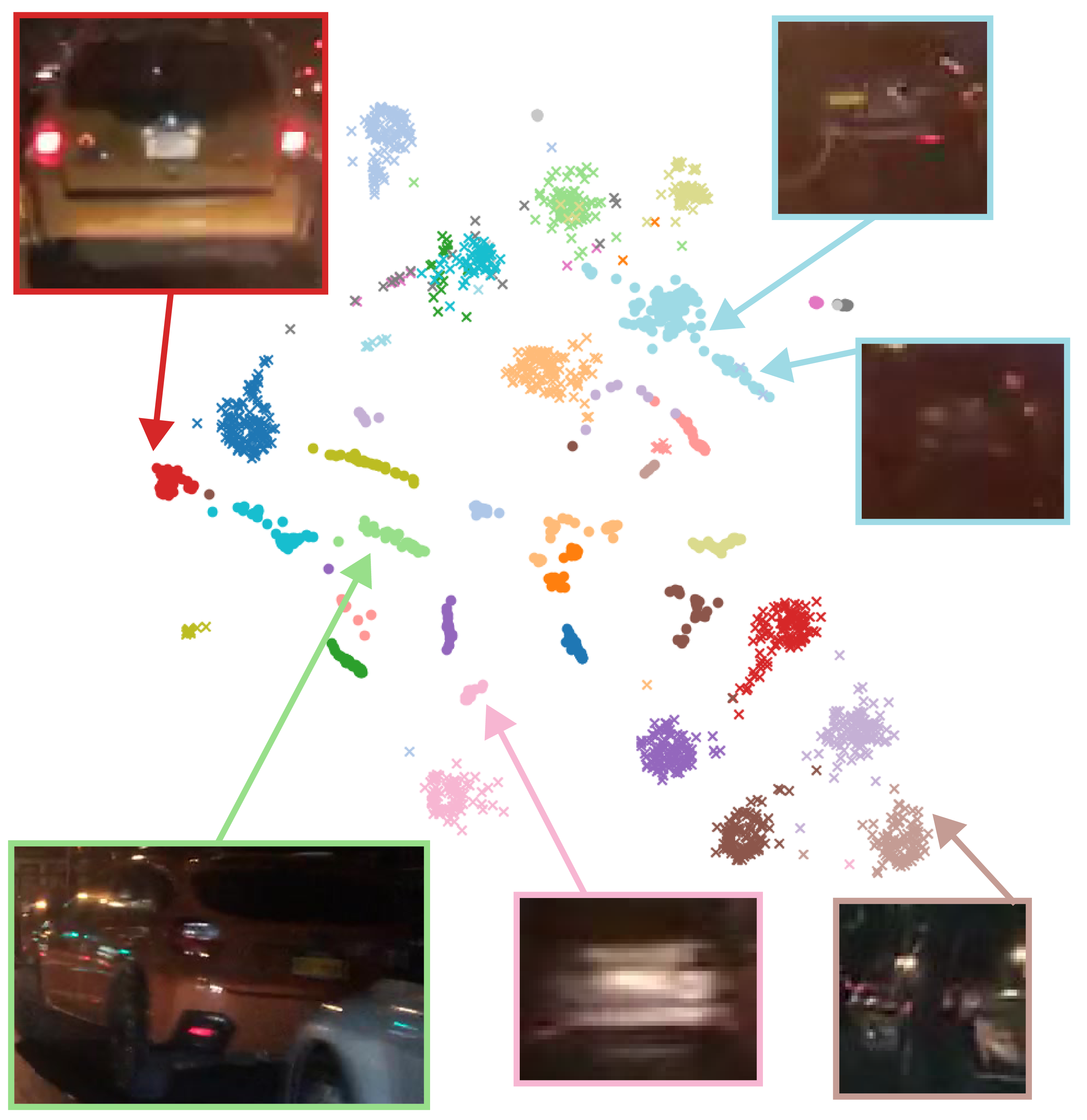}
    \caption{T-SNE projection of the predicted embeddings for the first 40 ground-truth objects in video \textit{b23f7012-fab06dac} of the BDD100K validation set. 
    Each color-symbol pair represents a ground-truth tracking ID assigned with DETR’s bipartite matching. Even during nighttime, the method can discriminate similar objects.}
    \label{fig:cover-illustration}
    \vspace{-0.7em}
\end{figure}

Likewise, the DEtection TRransformer (DETR) \cite{carion2020end} models object detection as a translation task, converting image features into object-level representations. 
Its cross-attention mechanism enables extracting relevant features for detection without the need for anchors, hand-crafted feature extraction methods, or local prediction biases inherent in CNN architectures \cite{girshick2014rcnn,ren2015fasterrcnn,redmon2016yolo}. The self-attention mechanism, coupled with bipartite matching \cite{kuhn1955hungarian,munkres1957algorithmshung}, helps object representations reach a consensus and eliminate redundancy in predictions.

Multi-Object Tracking (\ac{mot}) is a task that requires both object detection and object association. 
Object association links detected objects across frames by considering their appearance, position, size, or other characteristics to determine which detections correspond to the same real-world object. 
The algorithm must be robust against (partial) occlusion, loss of sight, missed detection, appearance variation, etc. In particular, the re-identification of objects after occlusions can be challenging. 

The object-level representations from DETR could be potentially used to re-identify objects across different frames. However, while each of these embeddings corresponds to a particular object in a specific frame, they do not provide a fine-grained description of the target appearance as needed for MOT, as the model is only trained to regress the bounding box and to classify the detected object. As a result, they are insufficient for object re-identification.

Our model, named Contrastive TRansformer (\ourmodelname{}), takes advantage of the representations produced by DETR to encode discriminative identity-level features for seamless object re-identification. This can be accomplished with little overhead using an instance-level contrastive loss and a revised sampling strategy. 
During training, the batch of images is built with sets of non-consecutive frames from multiple videos. It is important to select frames that are distant in time to increase the diversity of the appearance of objects, and to select images from several videos to have a wide diversity of negative examples for the contrastive loss. 
This sampling approach increases robustness under large object motion and adverse conditions, as illustrated in Figure \ref{fig:bdd100k_qualitative}.

Unlike competing DETR-like transformers \cite{meinhardt2022trackformer,sun2020transtrack,zeng2022motr,Zhang_2023_CVPR,korbar2022end,cai2022memot,zhu2022looking, 9964258}, we model \ac{mot} as a multi-task learning problem and introduce a model capable of performing joint detection and association with a single set of internal object representations. This simple yet effective approach removes the need for expensive additional transformer layers to update tracks \cite{sun2020transtrack,zeng2022motr,Zhang_2023_CVPR,zhu2022looking} or to perform object association \cite{cai2022memot,9964258}. 

Inspired by advances in supervised contrastive learning for image classification \cite{supervisedContrastive2020}, we propose a supplementary pre-training strategy on object detection datasets that does not require annotated tracking instance IDs and videos.
This method leverages the size and diversity of object detection datasets. We show that it improves the tracking embedding space and boosts performance even when fine-tuning on large tracking datasets. We evaluate our method on the MOT17 \cite{MOT16} benchmark and the more diverse and challenging BDD100K \cite{bdd100k} dataset.

Our main contributions can be summarized as follows:
\begin{itemize}[noitemsep]
    \item We show how to turn a DETR-like object detection model into a tracking model with little overhead by learning to discriminate instances. Our approach reaches performance on par or higher than more complicated architectures. Furthermore, it improves SOTA by + 2.6 mMOTA on the BDD100K dataset.
    \item We highlight the required components for learning robust joint detection and association features through an extensive ablation study.
    \item We present an additional pre-training scheme for MOT that takes advantage of the size and diversity of object detection datasets and is scalable to large object detection datasets.
\end{itemize}

%% file: related_work.tex
\begin{figure*}[t!]
\centering
    \begin{subfigure}[t]{0.31\textwidth}
        \centering
        \includegraphics[width=0.95\textwidth,trim={90px 80px 90px 80px},clip]{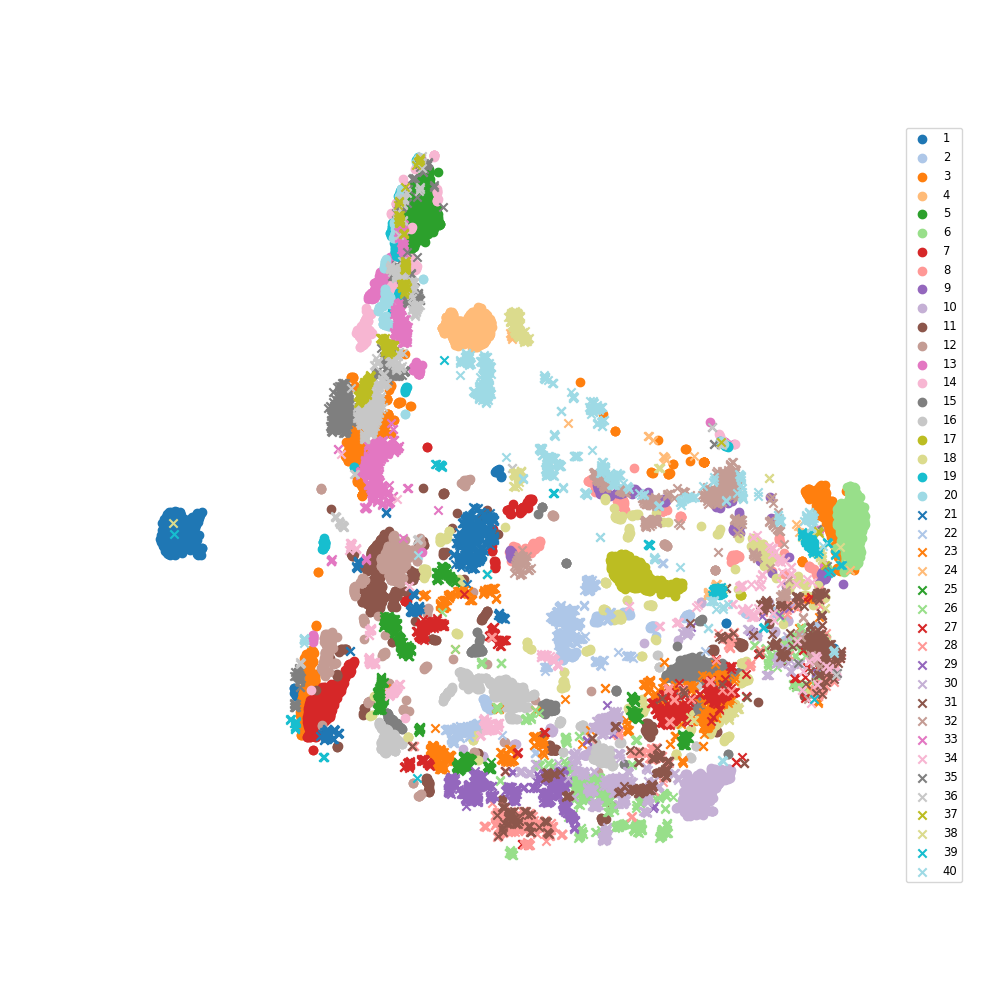}
        \caption{Pre-trained on detection without a contrastive loss. In this case, tracking embeddings correspond to the object embeddings.}
        \label{fig:tsne_mot17_ch_no-contrast-det}
    \end{subfigure}
    \hfill
    \begin{subfigure}[t]{0.31\textwidth}
        \centering
        \includegraphics[width=0.95\textwidth,trim={90px 80px 80px 80px},clip]{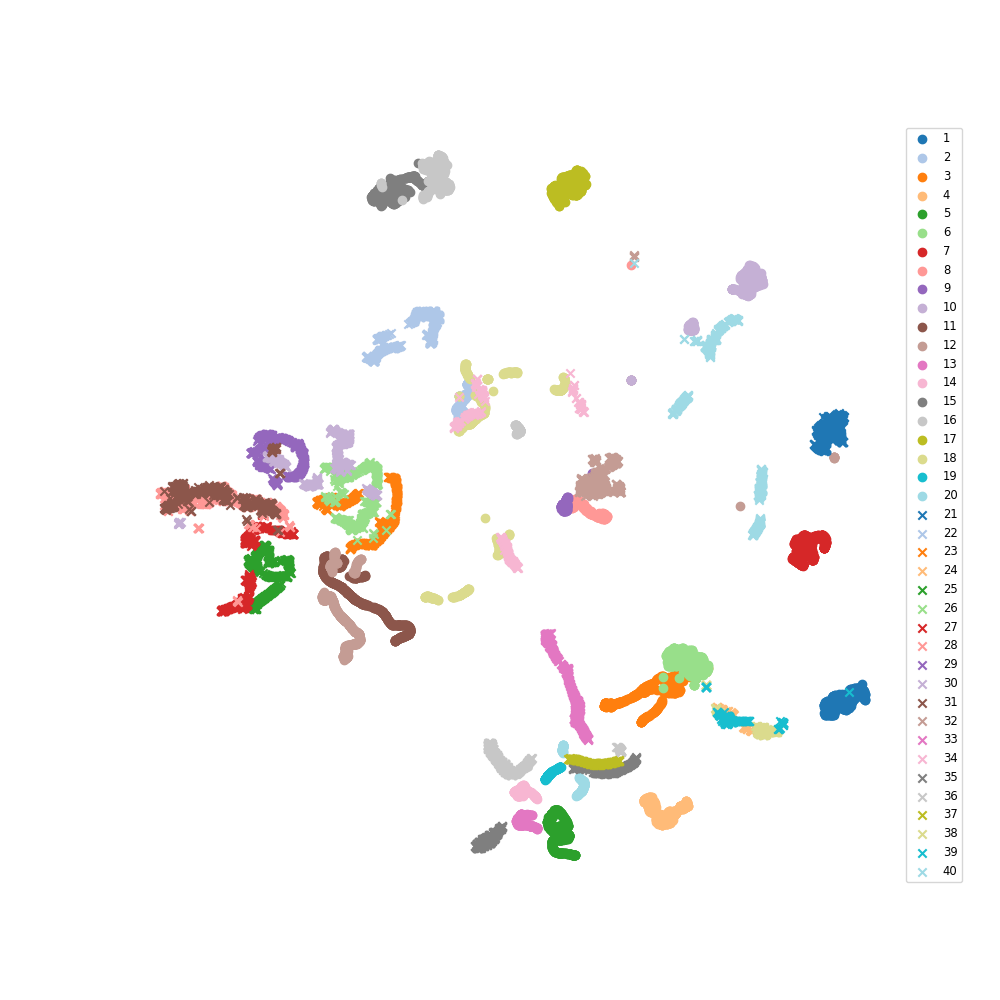}
        \caption{Pre-trained on detection with our contrastive loss (no tracking ID annotations required).}
        \label{fig:tsne_mot17_ch_with-contrast-det}
    \end{subfigure}
    \hfill
    \begin{subfigure}[t]{0.31\textwidth}
        \centering
        \includegraphics[width=0.95\textwidth,trim={90px 80px 70px 80px},clip]{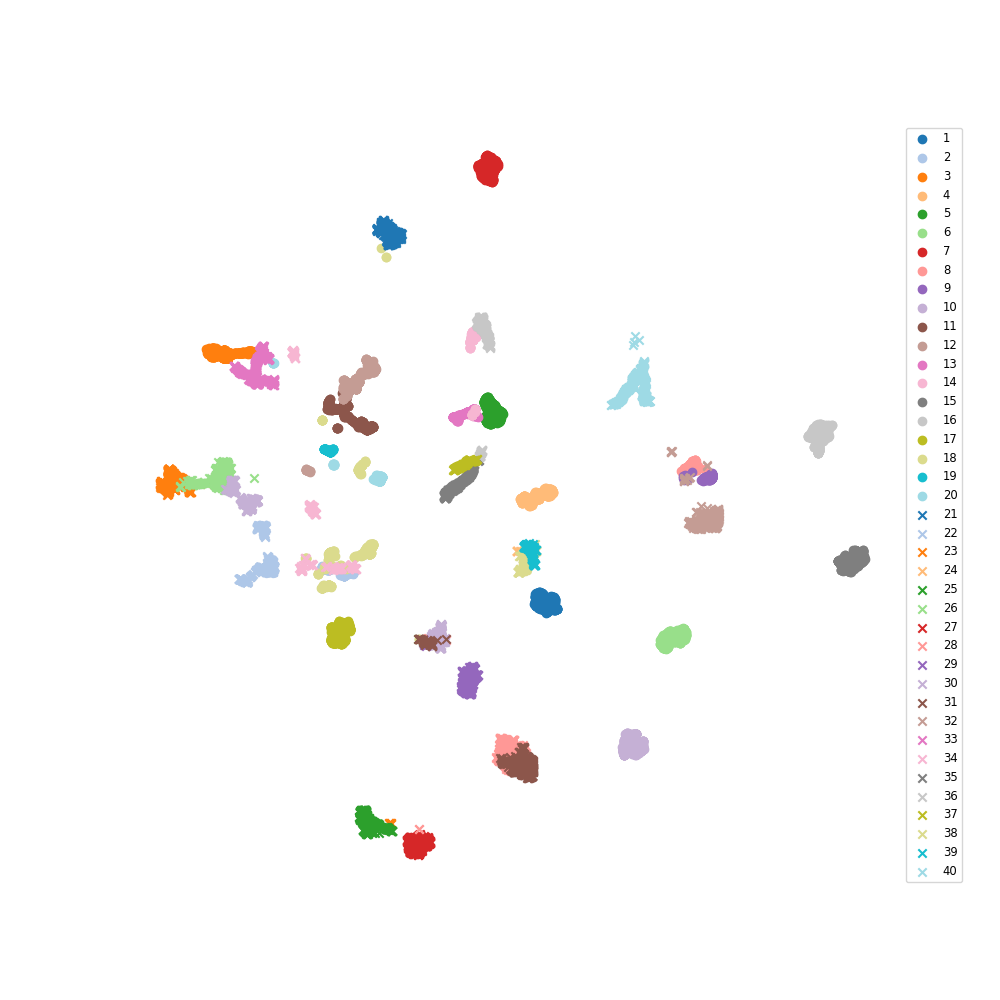}
        \caption{Pre-trained on detection with our contrastive loss, then trained for tracking on MOT17 with a contrastive loss.}
        \label{fig:tsne_mot17_with-tracking-phase}
    \end{subfigure}
    \caption{t-SNE visualization of the tracking embeddings of video 4 of MOT17. Each color-symbol pair represents a unique tracking ID, assigned with DETR's bipartite matching. All models are pre-trained on the CrowdHuman dataset \cite{shao2018crowdhuman} and evaluated on the validation set of MOT17 \cite{MOT16}. Without the contrastive loss, Deformable-DETR's embeddings are not clustered per instance id.}
    \label{fig:embeddings-tsne}
    \vspace{-0.5em}
\end{figure*}

\section{Related Work}

This section briefly overviews joint detection and tracking methods, contrastive learning, and the use of DETR-like transformers for MOT.

\paragrax{Joint-detection-and-tracking.} Several works are focused on the implementation of models capable of performing detection and tracking \cite{wang2020JDE_towards, zhang2021fairmot, lu2020retinatrack, cvpr_qdtrack, Feichtenhofer_2017_ICCV, zhou2020tracking, tokmakov2021learning, tsai2023swinJDE}. In particular, some of them \cite{wang2020JDE_towards, zhang2021fairmot, lu2020retinatrack, cvpr_qdtrack, tsai2023swinJDE} train the detection model to extract appearance embeddings used to re-identify objects across frames.

JDE \cite{wang2020JDE_towards} introduced a CNN-based joint detection and tracking model that prioritizes speed. Specifically, a detection model is trained to generate discriminative feature embeddings using a cross-entropy loss with one category per unique instance.
FairMOT \cite{zhang2021fairmot} presented a set of design improvements to mitigate the negative effect of anchors and high dimensional re-ID features when training a joint detection and tracking model. Furthermore, FairMOT introduced a scheme for training a tracker using individual images from detection datasets, bypassing the requirement for sequences. 
However, the scalability of their cross-entropy loss-based training procedure is limited for massive datasets like BDD100K, requiring a classification head with a significant number of parameters (approximately 230 million for the projection head of FairMOT on the BDD100K detection set). In contrast, our training approach decouples the tracking model size from the dataset size, ensuring scalability for large-scale datasets.

\paragrax{Contrastive Learning.} Self-supervised contrastive learning has become a popular technique for learning robust and discriminative feature representations in computer vision \cite{chen2020simple,henaff2020data,hjelmlearning,tian2020contrastive, he2020momentum}. For example, \cite{chen2020simple} learns representations by augmenting each image in a batch twice and encouraging the similarity between views of the same image while encouraging dissimilarity with views of different images. 
Later, supervised contrastive learning for image classification \cite{supervisedContrastive2020} has been proposed as a more effective method for learning feature representations, where labeled data is used to guide the contrastive loss function. In addition to the augmented views, images from the same class are also considered positive examples. This improves classification accuracy by a significant margin.

Contrastive learning also finds applications in tracking. QDTrack \cite{cvpr_qdtrack} learns appearance with a contrastive loss on object regions by sampling pairs of images from neighboring frames, whereas we sample multiple frames per video and from different videos, allowing for a much larger diversity of positive and negative examples for contrastive learning. MTrack \cite{yu2022towards} predicts a set of feature vectors for each object in an image. Contrastive learning is then used to push an object's inter-frame and intra-frame feature vectors to be similar and feature vectors from other objects to be dissimilar.
Since multiple feature vectors are generated per object and frame, directly using contrastive learning would be costly. MTrack instead pushes feature vectors towards their corresponding trajectory centers.

\paragrax{Transformers for Multi-Object Tracking.}  
DETR \cite{carion2020end} introduced a simple one-stage framework for object detection, removing the need to create anchor boxes manually and for Non-Maximum Suppression (NMS). 
It models object detection as a sequence-to-sequence translation task: it first extracts image features, which then serve as keys and values to update a set of learnable object queries through a series of transformer decoder blocks. 
Multiple works have built trackers on top of DETR by exploiting the object-level internal representations for class-agnostic MOT \cite{korbar2022end} and MOT \cite{meinhardt2022trackformer,zeng2022motr,sun2020transtrack,cai2022memot,zhu2022looking, 9964258}. 

TrackFormer \cite{meinhardt2022trackformer} and MOTR \cite{zeng2022motr} introduced track queries to model instances over time and preserve identities.
Newborn objects are detected with object queries, and their hidden states produce track queries for the next frame. Both query types are fed into the transformer decoder.
TrackFormer is trained on consecutive frame pairs, which prevents its application for long-range occlusions. MOTR instead learns long-range dependencies through a cross-attention temporal aggregation network and a video-level loss. 
Both approaches require track augmentations and face challenges in detecting newborn objects. 
MOTRv2 \cite{Zhang_2023_CVPR} solved the poor performance on newborn objects by using an additional YOLOX \cite{ge2021yolox} as a proposal network, but at the cost of running this additional detection model.

Similarly, MeMOT \cite{cai2022memot} uses a memory aggregation module to encode temporal object information from a memory of previous embeddings into track embeddings through attention modules. 
The model is then trained to predict uniqueness scores so that only new objects are added to the track embeddings.

TransTrack \cite{sun2020transtrack} instead uses a unique set of object queries to detect objects. 
Additionally, a parallel transformer decoder autoregressively updates tracked objects. 
The association is performed by matching the outputs of both decoders with an IoU cost,
which leads to poor association performance. In contrast, our approach avoids a parallel decoder, opting for learning instance-level embeddings that are more robust for re-identification.

%% file: model.tex
\section{Learning Identity-Level Representations}

We introduce a method that can turn any DETR-like object detector into a model that can perform joint detection and \ac{mot}. 
We describe how such a model can learn identity-level features with the aid of an additional loss term and a revisited sampling strategy. We further introduce a pre-training scheme that allows us to learn better representations by exploiting large-scale object detection datasets.

\subsection{Preliminaries}
Given an input image $I$, DETR extracts image features with a backbone and a transformer encoder.
These features then serve as keys and values to update a set of $N$ learnable object queries through a series of transformer decoder blocks. 
Each output object embedding $\left\{\hat{x}_1^I, \hat{x}_2^I, \cdots, \hat{x}_N^I\right\}$ represents a different possible object in the image.
Class probabilities and bounding box positions are then predicted for each output embedding through \ac{ffn} heads.

The method uses the Hungarian algorithm \cite{kuhn1955hungarian,munkres1957algorithmshung} to match ground truth objects with predictions, forming an optimal bipartite matching $\hat{\sigma}^I$. The assignment minimizes the global matching cost, which is a linear combination of classification and localization costs. 
The matched predictions are then improved with object-specific losses (i.e. classification and localization losses), and the remaining predictions are trained to predict the background class. 

\subsection{\ac{mot} as a multi-task learning problem}

Our method exploits the object-level representations from DETR for re-identification.
As illustrated in Figure \ref{fig:tsne_mot17_ch_no-contrast-det}, these representations do not provide a fine-grained description of the objects' appearance. We, therefore, introduce an additional loss term to force the model to encode identity-level features besides the bounding box and class information. 
More specifically, we build an embedding space for tracking in which different views of an object are positioned close to one another, whereas other objects lie further in the embedding space, see Figure \ref{fig:tsne_mot17_with-tracking-phase}.

\paragrax{Tracking projection head.} 
We pass each output embedding of the decoder to three shared \ac{ffn}s that predict respectively class probabilities, bounding box positions and \emph{tracking embeddings} $\left\{\hat{z}_1^I, \hat{z}_2^I, \cdots, \hat{z}_N^I\right\}$.
The projection into \emph{tracking embeddings} is more convenient for performing object re-identification.

\paragrax{Contrastive learning.} Our framework relies on the bipartite matching $\hat{\sigma}^I$ to associate output object embeddings to ground-truths objects and, therefore, the corresponding tracking embeddings to the annotated tracking IDs. As a result, each tracking embedding $\hat{z}_i^I \in \mathbb{R}^D$ that was associated with a ground truth will be associated with an object instance ID: $\left\{(\hat{z}_{\hat{\sigma}^I(1)}^I, ID_1^I), (\hat{z}_{\hat{\sigma}^I(2)}^I, ID_2^I), \cdots,  (\hat{z}_{\hat{\sigma}^I(M)}^I, ID_M^I)\right\}$ where $M$ is the number of ground truth objects in the image.
For simplicity of notations, the set of tracking embeddings matched with an object over the full batch is denoted by:
\begin{equation}
\begin{aligned}[b]
    \tilde{\bm{z}} 
        & = \left[ \hat{z}_{\hat{\sigma}^I(1)}^1, \hat{z}_{\hat{\sigma}^I(2)}^1, \cdots, \hat{z}_{{\hat{\sigma}^I(M_1)}}^1, \hat{z}_{\hat{\sigma}^I(1)}^2, \cdots, \hat{z}_{\hat{\sigma}^I{(M_B)}}^B \right]
\end{aligned}
\end{equation}
where $M_{b}$ is the total number of ground truth objects in video $b$ and $B$ is the batch size.

Given a tracking embedding $\tilde{z}_{i}$ and its corresponding ground truth instance id $ID_{i}$, we define the set of positives $\mathcal{P}(i)$ as embeddings in the batch that are associated with objects from the same video with the same ground truth tracking id $ID_i$. The remaining are negatives $\mathcal{N}(i)$. 
Following previous work on supervised contrastive learning \cite{supervisedContrastive2020}, the contrastive loss for tracking embeddings $\tilde{z}_{i}$ and $\tilde{z}_{j}$ corresponding to two views from the same instance (i.e. such that $ID_{i} = ID_{j}$) takes the following form.
\begin{multline}
    l_{\mathrm{contr}}\left(\tilde{z}_{i}, \tilde{z}_{j}, \mathcal{N}(i) \right) = \\
    - \log\left( \frac{e^{\textit{sim}(\tilde{z}_{i}, \tilde{z}_{j}) / \tau}}{e^{\textit{sim}(\tilde{z}_{i}, \tilde{z}_{j}) / \tau} + \sum_{k \in \mathcal{N}(i) }e^{\textit{sim}(\tilde{z}_{i}, \tilde{z}_{k}) / \tau}} \right)
    \label{eq:lcontr}
\end{multline}
where $\tau \in \mathbb{R}^{+}$ is the temperature hyper-parameter.

Note that only the similarity of the target positive pair $(i, j)$ and of the negative examples $\mathcal{N}(i)$ is considered in the denominator so that the loss value for a given pair is invariant to the other positive pairs.  
The contrastive loss is then averaged over the set of positive pairs:
\begin{equation}
    \mathcal{L}_{\mathrm{contr}}\left(\tilde{z}_i, \mathcal{P}(i), \mathcal{N}(i) \right) = \frac{1}{ \left| \mathcal{P}(i) \right| } \sum_{j \in \mathcal{P}(i)} l_{\mathrm{contr}}\left(\tilde{z}_{i}, \tilde{z}_{j}, \mathcal{N}(i) \right)
\end{equation}

where $\left| \mathcal{P}(i) \right|$ denotes the number of positives for ground truth object $i$ in the batch. 

Such a loss forces the tracking embeddings associated with the same identity to be pulled as close as possible while tracking embeddings of different identities to be pushed as far as possible. 

We then reformulate the overall loss as a weighted sum of per-object classification, localization, and contrastive loss terms:
\begin{equation}
\begin{aligned}
    \mathcal{L}_{\mathrm{train}}(\bm{y}_i, \hat{\bm{y}}_{i}) =
    & \lambda_{\mathrm{class}} \mathcal{L}_{\mathrm{class}}(c_i, \hat{c}_{\hat{\sigma}(i)}) 
    \\ & + \mathbbm{1}_{\left\{c_i \neq \varnothing\right\}} \lambda_{\mathrm{box}}\mathcal{L}_{\mathrm{box}}(b_i, \hat{b}_{\hat{\sigma}(i)})  
    \\ & + \mathbbm{1}_{\left\{c_i \neq \varnothing\right\}} \lambda_{\mathrm{contr}}\mathcal{L}_{\mathrm{contr}}\left(\tilde{z}_i, \mathcal{P}(i), \mathcal{N}(i) \right)
\end{aligned}
\label{eq:loss-function}
\end{equation}

where $\mathcal{L}_{\mathrm{class}}$, $\mathcal{L}_{\mathrm{box}}$ and $\mathcal{L}_{\mathrm{contr}}$ are respectively the classification, localization and contrastive loss terms with their corresponding weighting hyper-parameters $\lambda$, $c_i$ is the target class label, $b_i \in [0,1]^4$ is the target bounding box in coordinates relative to the image size.

\subsection{Sampling strategy}
The default data sampling strategy to build training batches for object detection is to sample images from the training set with a uniform distribution. This allows each mini-batch to be representative of the training set. However, such a way of sampling images dramatically reduces the probability of having multiple views of the same ground truth object in different images. This implies a very small, if not non-existent, set of positive pairs within each batch. Yet, contrastive learning usually benefits from having many positive pairs \cite{supervisedContrastive2020}. We, therefore, design an alternative sampling strategy with the following properties:

\begin{itemize}[noitemsep,topsep=0pt]
    \item a high probability of having many positives by sampling multiple frames from each video in the batch;
    \item variation in objects' appearance by sampling non-consecutive frames;
    \item a large diversity in the negative examples by sampling from multiple videos.
\end{itemize}

We build training batches by sampling with a uniform distribution $N_v$ videos and then sampling $N_f$ frames from each of these videos again with a uniform distribution. This sampling strategy increases the probability of including the same identities from different frames, potentially including more variations of the same object. At the same time, it also ensures a diverse enough selection of objects with different identities and contexts, as the frames also come from different videos. Besides, such a sampling strategy practically preserves the batch variety needed not to compromise the model's object detection capabilities.

\subsection{Learning MOT from Object Detection datasets}
\label{sec:learnmotfromdet}

Most tracking methods are pre-trained on object detection datasets, which are usually larger and more diverse than MOT datasets. 
However, MOT requires features to be informative enough to discriminate different instances of the same class, and object detectors don’t offer the granularity level required for object association. 
We show how to learn an embedding space that discriminates between instances during the object detection pre-training. We then empirically verify that this pre-training phase offers a better initialization for the MOT training phase.

We follow a similar approach to self-supervised contrastive learning on image classification \cite{chen2020simple,henaff2020data,hjelmlearning,tian2020contrastive}.
For a set of $N$ randomly sampled images, we build a batch of $2N$ images by applying two different sets of augmentations to each image. 
We then train the model with a supervised contrastive loss in which representations of the same ground truth object are pushed to be similar, whereas other object representations are pushed to be dissimilar.

\begin{figure}[t]
    \centering
    \def\svgwidth{0.94\columnwidth}
    {\fontsize{6pt}{6pt}\selectfont 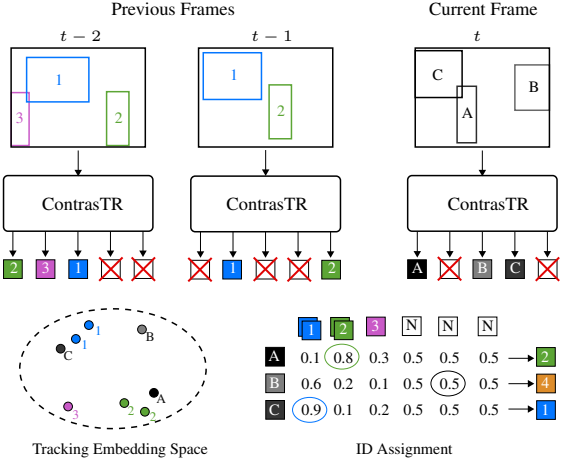}
    \caption{\footnotesize{\textbf{\ourmodelname{}.} Our framework during the \emph{inference} phase. ID assignment maximizes the global cosine similarity between the predictions and previous instances. A \textit{new instance} (N) entry is added to the cost matrix for each prediction.}}
    \label{fig:inference_contrasTR}
    \vspace{-0.9em}
\end{figure}

\subsection{Object association with maximal similarity} 

At inference time, frames are processed online. Detections with a confidence score above the \textit{objectness threshold} are collected into a memory queue $m$ that allows up to $T$ most recent frames in the past. It works essentially as a first-in-first-out (FIFO) queue. 
This memory allows the model to re-identify objects even after occlusions. 
The ID assignment process is the following:

\begin{enumerate}[noitemsep]
    \item The model generates object embeddings for the current frame and uses the projection head to generate tracking embeddings for each non-background prediction. 
    \item A similarity matrix is formed with new predictions as rows and previously predicted instances as columns. The embeddings from previous frames are grouped per instance, where the highest similarity value is kept.
    \begin{equation}
        s_{i,j} = \max_{t^\prime \in [t-T-1, t-1]} \textit{sim}(\hat{z}_{i}, m_{j}^{t^\prime})
    \end{equation}
    where $m_{j}^{t^\prime}$ is the memory of tracking embedding with assigned id $j$ at frame $t^\prime$.
    An additional \textit{new instance} entry is added to the cost matrix for each predicted embedding, with a pre-defined \textit{tracking threshold} score.
    \item The Hungarian algorithm finds a bipartite matching between the predicted and the previous tracking embeddings.  
    The solution maximizes the sum of the similarities of the assignment.
    \begin{equation}
        \hat{\mu} = \underset{\mu \in \mathcal{M}}{\arg \max } \sum_{i=1}^{K}  s_{i,\mu(i)} 
        \label{eq:matching}
    \end{equation} 
    where $K$ is the number of non-background predictions and $\mathcal{M}$ is the set of all possible assignments. 
    \item Each object is assigned the ID of the corresponding previous embedding. If an object is matched with a \textit{new instance} entry, it is assigned a new unique ID.
\end{enumerate}
The process is repeated frame-by-frame online and is illustrated on Figure \ref{fig:inference_contrasTR}.

It should be noted that the memory size $T$ can be adjusted during inference time, based on the specific requirements of the target application and the video frame rate, without the need to re-train the model.

%% file: ContrasTR_inference.pdf_tex
\begingroup%
  \makeatletter%
  \providecommand\color[2][]{%
    \errmessage{(Inkscape) Color is used for the text in Inkscape, but the package 'color.sty' is not loaded}%
    \renewcommand\color[2][]{}%
  }%
  \providecommand\transparent[1]{%
    \errmessage{(Inkscape) Transparency is used (non-zero) for the text in Inkscape, but the package 'transparent.sty' is not loaded}%
    \renewcommand\transparent[1]{}%
  }%
  \providecommand\rotatebox[2]{#2}%
  \newcommand*\fsize{\dimexpr\f@size pt\relax}%
  \newcommand*\lineheight[1]{\fontsize{\fsize}{#1\fsize}\selectfont}%
  \ifx\svgwidth\undefined%
    \setlength{\unitlength}{880.84011154bp}%
    \ifx\svgscale\undefined%
      \relax%
    \else%
      \setlength{\unitlength}{\unitlength * \real{\svgscale}}%
    \fi%
  \else%
    \setlength{\unitlength}{\svgwidth}%
  \fi%
  \global\let\svgwidth\undefined%
  \global\let\svgscale\undefined%
  \makeatother%
  \begin{picture}(1,0.77368266)%
    \lineheight{1}%
    \setlength\tabcolsep{0pt}%
    \put(0,0){\includegraphics[width=\unitlength,page=1]{ContrasTR_inference.pdf}}%
    \put(0.06695029,0.42107437){\makebox(0,0)[lt]{\lineheight{1.25}\smash{\begin{tabular}[t]{l}{\scriptsize  ContrasTR}\end{tabular}}}}%
    \put(0,0){\includegraphics[width=\unitlength,page=2]{ContrasTR_inference.pdf}}%
    \put(0.08872212,0.63330075){\color[rgb]{0.01176471,0.4627451,1}\makebox(0,0)[lt]{\lineheight{1.25}\smash{\begin{tabular}[t]{l}1\end{tabular}}}}%
    \put(0.18989258,0.56855574){\color[rgb]{0.36078431,0.64705882,0.20784314}\makebox(0,0)[lt]{\lineheight{1.25}\smash{\begin{tabular}[t]{l}2\end{tabular}}}}%
    \put(0.12176727,0.31291343){\color[rgb]{1,1,1}\makebox(0,0)[lt]{\lineheight{1.25}\smash{\begin{tabular}[t]{l}1\end{tabular}}}}%
    \put(0.01243983,0.31291343){\color[rgb]{1,1,1}\makebox(0,0)[lt]{\lineheight{1.25}\smash{\begin{tabular}[t]{l}2\end{tabular}}}}%
    \put(0.02218053,0.56855574){\color[rgb]{0.78823529,0.34901961,0.78039216}\makebox(0,0)[lt]{\lineheight{1.25}\smash{\begin{tabular}[t]{l}3\end{tabular}}}}%
    \put(0.06609883,0.31258987){\color[rgb]{1,1,1}\makebox(0,0)[lt]{\lineheight{1.25}\smash{\begin{tabular}[t]{l}3\end{tabular}}}}%
    \put(0.09657258,0.70813555){\makebox(0,0)[lt]{\lineheight{1.25}\smash{\begin{tabular}[t]{l}$t-2$\end{tabular}}}}%
    \put(0,0){\includegraphics[width=\unitlength,page=3]{ContrasTR_inference.pdf}}%
    \put(0.37970853,0.42107437){\makebox(0,0)[lt]{\lineheight{1.25}\smash{\begin{tabular}[t]{l}{\scriptsize  ContrasTR}\end{tabular}}}}%
    \put(0.75543222,0.42107437){\makebox(0,0)[lt]{\lineheight{1.25}\smash{\begin{tabular}[t]{l}{\scriptsize  ContrasTR}\end{tabular}}}}%
    \put(0,0){\includegraphics[width=\unitlength,page=4]{ContrasTR_inference.pdf}}%
    \put(0.42189836,0.70813555){\makebox(0,0)[lt]{\lineheight{1.25}\smash{\begin{tabular}[t]{l}$t-1$\end{tabular}}}}%
    \put(0,0){\includegraphics[width=\unitlength,page=5]{ContrasTR_inference.pdf}}%
    \put(0.38601784,0.64167911){\color[rgb]{0.01176471,0.4627451,1}\makebox(0,0)[lt]{\lineheight{1.25}\smash{\begin{tabular}[t]{l}1\end{tabular}}}}%
    \put(0.46568894,0.58010153){\color[rgb]{0.36078431,0.64705882,0.20784314}\makebox(0,0)[lt]{\lineheight{1.25}\smash{\begin{tabular}[t]{l}2\end{tabular}}}}%
    \put(0.3840084,0.31258987){\color[rgb]{1,1,1}\makebox(0,0)[lt]{\lineheight{1.25}\smash{\begin{tabular}[t]{l}1\end{tabular}}}}%
    \put(0.55197588,0.31258987){\color[rgb]{1,1,1}\makebox(0,0)[lt]{\lineheight{1.25}\smash{\begin{tabular}[t]{l}2\end{tabular}}}}%
    \put(0,0){\includegraphics[width=\unitlength,page=6]{ContrasTR_inference.pdf}}%
    \put(0.18450284,0.75168772){\makebox(0,0)[lt]{\lineheight{1.25}\smash{\begin{tabular}[t]{l}{\scriptsize  Previous  Frames}\end{tabular}}}}%
    \put(0.72365574,0.75105764){\makebox(0,0)[lt]{\lineheight{1.25}\smash{\begin{tabular}[t]{l}{\scriptsize  Current  Frame}\end{tabular}}}}%
    \put(0.79890776,0.70813555){\makebox(0,0)[lt]{\lineheight{1.25}\smash{\begin{tabular}[t]{l}$t$\end{tabular}}}}%
    \put(0,0){\includegraphics[width=\unitlength,page=7]{ContrasTR_inference.pdf}}%
    \put(0.7292328,0.64167911){\makebox(0,0)[lt]{\lineheight{1.25}\smash{\begin{tabular}[t]{l}C\end{tabular}}}}%
    \put(0.89284081,0.62086943){\makebox(0,0)[lt]{\lineheight{1.25}\smash{\begin{tabular}[t]{l}B\end{tabular}}}}%
    \put(0.77907726,0.57674678){\makebox(0,0)[lt]{\lineheight{1.25}\smash{\begin{tabular}[t]{l}A\end{tabular}}}}%
    \put(0,0){\includegraphics[width=\unitlength,page=8]{ContrasTR_inference.pdf}}%
    \put(0.69420374,0.31297303){\color[rgb]{1,1,1}\makebox(0,0)[lt]{\lineheight{1.25}\smash{\begin{tabular}[t]{l}A\end{tabular}}}}%
    \put(0.80578756,0.31297303){\color[rgb]{1,1,1}\makebox(0,0)[lt]{\lineheight{1.25}\smash{\begin{tabular}[t]{l}B\end{tabular}}}}%
    \put(0.86083444,0.31285383){\color[rgb]{1,1,1}\makebox(0,0)[lt]{\lineheight{1.25}\smash{\begin{tabular}[t]{l}C\end{tabular}}}}%
    \put(0,0){\includegraphics[width=\unitlength,page=9]{ContrasTR_inference.pdf}}%
    \put(0.11808045,0.06514712){\color[rgb]{0.78823529,0.34901961,0.78039216}\makebox(0,0)[lt]{\lineheight{1.25}\smash{\begin{tabular}[t]{l}{\tiny 3}\end{tabular}}}}%
    \put(0,0){\includegraphics[width=\unitlength,page=10]{ContrasTR_inference.pdf}}%
    \put(0.05023613,0.00602175){\makebox(0,0)[lt]{\lineheight{1.25}\smash{\begin{tabular}[t]{l}Tracking Embedding Space\end{tabular}}}}%
    \put(0.24281932,0.19757468){\color[rgb]{0.05882353,0.05882353,0.05882353}\makebox(0,0)[lt]{\lineheight{1.25}\smash{\begin{tabular}[t]{l}{\tiny B}\end{tabular}}}}%
    \put(0.15684742,0.2087288){\color[rgb]{0,0.47058824,1}\makebox(0,0)[lt]{\lineheight{1.25}\smash{\begin{tabular}[t]{l}{\tiny 1}\end{tabular}}}}%
    \put(0.21243924,0.07276769){\color[rgb]{0.36078431,0.64705882,0.20784314}\makebox(0,0)[lt]{\lineheight{1.25}\smash{\begin{tabular}[t]{l}{\tiny 2}\end{tabular}}}}%
    \put(0.24949477,0.06199672){\color[rgb]{0.36078431,0.64705882,0.20784314}\makebox(0,0)[lt]{\lineheight{1.25}\smash{\begin{tabular}[t]{l}{\tiny 2}\end{tabular}}}}%
    \put(0.13353445,0.18474318){\color[rgb]{0,0.47058824,1}\makebox(0,0)[lt]{\lineheight{1.25}\smash{\begin{tabular}[t]{l}{\tiny 1}\end{tabular}}}}%
    \put(0.10379296,0.16504891){\color[rgb]{0.05882353,0.05882353,0.05882353}\makebox(0,0)[lt]{\lineheight{1.25}\smash{\begin{tabular}[t]{l}{\tiny C }\end{tabular}}}}%
    \put(0.26071701,0.0924279){\color[rgb]{0.05882353,0.05882353,0.05882353}\makebox(0,0)[lt]{\lineheight{1.25}\smash{\begin{tabular}[t]{l}{\tiny A }\end{tabular}}}}%
    \put(0,0){\includegraphics[width=\unitlength,page=11]{ContrasTR_inference.pdf}}%
    \put(0.50594596,0.16113219){\makebox(0,0)[lt]{\lineheight{1.25}\smash{\begin{tabular}[t]{l}0.1\end{tabular}}}}%
    \put(0.56151224,0.16104705){\makebox(0,0)[lt]{\lineheight{1.25}\smash{\begin{tabular}[t]{l}0.8\end{tabular}}}}%
    \put(0.62237459,0.16042548){\makebox(0,0)[lt]{\lineheight{1.25}\smash{\begin{tabular}[t]{l}0.3\end{tabular}}}}%
    \put(0.50687406,0.1165157){\makebox(0,0)[lt]{\lineheight{1.25}\smash{\begin{tabular}[t]{l}0.6\end{tabular}}}}%
    \put(0.56151224,0.11666044){\makebox(0,0)[lt]{\lineheight{1.25}\smash{\begin{tabular}[t]{l}0.2\end{tabular}}}}%
    \put(0.62237459,0.11603888){\makebox(0,0)[lt]{\lineheight{1.25}\smash{\begin{tabular}[t]{l}0.1\end{tabular}}}}%
    \put(0.50687406,0.07212909){\makebox(0,0)[lt]{\lineheight{1.25}\smash{\begin{tabular}[t]{l}0.9\end{tabular}}}}%
    \put(0.56151224,0.07226533){\makebox(0,0)[lt]{\lineheight{1.25}\smash{\begin{tabular}[t]{l}0.1\end{tabular}}}}%
    \put(0.62237459,0.07164376){\makebox(0,0)[lt]{\lineheight{1.25}\smash{\begin{tabular}[t]{l}0.2\end{tabular}}}}%
    \put(0.68012911,0.16091081){\makebox(0,0)[lt]{\lineheight{1.25}\smash{\begin{tabular}[t]{l}0.5\end{tabular}}}}%
    \put(0.68012911,0.11638798){\makebox(0,0)[lt]{\lineheight{1.25}\smash{\begin{tabular}[t]{l}0.5\end{tabular}}}}%
    \put(0.68012911,0.07183108){\makebox(0,0)[lt]{\lineheight{1.25}\smash{\begin{tabular}[t]{l}0.5\end{tabular}}}}%
    \put(0,0){\includegraphics[width=\unitlength,page=12]{ContrasTR_inference.pdf}}%
    \put(0.5997513,0.00602175){\makebox(0,0)[lt]{\lineheight{1.25}\smash{\begin{tabular}[t]{l}ID  Assignment\end{tabular}}}}%
    \put(0.45113749,0.16315015){\color[rgb]{1,1,1}\makebox(0,0)[lt]{\lineheight{1.25}\smash{\begin{tabular}[t]{l}A\end{tabular}}}}%
    \put(0.45252537,0.11802278){\color[rgb]{1,1,1}\makebox(0,0)[lt]{\lineheight{1.25}\smash{\begin{tabular}[t]{l}B\end{tabular}}}}%
    \put(0.4521337,0.07276769){\color[rgb]{1,1,1}\makebox(0,0)[lt]{\lineheight{1.25}\smash{\begin{tabular}[t]{l}C\end{tabular}}}}%
    \put(0,0){\includegraphics[width=\unitlength,page=13]{ContrasTR_inference.pdf}}%
    \put(0.91349723,0.16240938){\color[rgb]{1,1,1}\makebox(0,0)[lt]{\lineheight{1.25}\smash{\begin{tabular}[t]{l}2\end{tabular}}}}%
    \put(0.9141784,0.11728201){\color[rgb]{1,1,1}\makebox(0,0)[lt]{\lineheight{1.25}\smash{\begin{tabular}[t]{l}4\end{tabular}}}}%
    \put(0.9141784,0.07202692){\color[rgb]{1,1,1}\makebox(0,0)[lt]{\lineheight{1.25}\smash{\begin{tabular}[t]{l}1\end{tabular}}}}%
    \put(0.57154243,0.20921413){\color[rgb]{1,1,1}\makebox(0,0)[lt]{\lineheight{1.25}\smash{\begin{tabular}[t]{l}2\end{tabular}}}}%
    \put(0.51690425,0.20834564){\color[rgb]{1,1,1}\makebox(0,0)[lt]{\lineheight{1.25}\smash{\begin{tabular}[t]{l}1\end{tabular}}}}%
    \put(0,0){\includegraphics[width=\unitlength,page=14]{ContrasTR_inference.pdf}}%
    \put(0.6259422,0.21479971){\color[rgb]{1,1,1}\makebox(0,0)[lt]{\lineheight{1.25}\smash{\begin{tabular}[t]{l}3\end{tabular}}}}%
    \put(0.6845567,0.21587255){\makebox(0,0)[lt]{\lineheight{1.25}\smash{\begin{tabular}[t]{l}N\end{tabular}}}}%
    \put(0.7417152,0.16057023){\makebox(0,0)[lt]{\lineheight{1.25}\smash{\begin{tabular}[t]{l}0.5\end{tabular}}}}%
    \put(0.7417152,0.11604739){\makebox(0,0)[lt]{\lineheight{1.25}\smash{\begin{tabular}[t]{l}0.5\end{tabular}}}}%
    \put(0.7417152,0.07149901){\makebox(0,0)[lt]{\lineheight{1.25}\smash{\begin{tabular}[t]{l}0.5\end{tabular}}}}%
    \put(0,0){\includegraphics[width=\unitlength,page=15]{ContrasTR_inference.pdf}}%
    \put(0.74615131,0.21554048){\makebox(0,0)[lt]{\lineheight{1.25}\smash{\begin{tabular}[t]{l}N\end{tabular}}}}%
    \put(0.80881024,0.16057023){\makebox(0,0)[lt]{\lineheight{1.25}\smash{\begin{tabular}[t]{l}0.5\end{tabular}}}}%
    \put(0.80881024,0.11604739){\makebox(0,0)[lt]{\lineheight{1.25}\smash{\begin{tabular}[t]{l}0.5\end{tabular}}}}%
    \put(0.80881024,0.07149901){\makebox(0,0)[lt]{\lineheight{1.25}\smash{\begin{tabular}[t]{l}0.5\end{tabular}}}}%
    \put(0,0){\includegraphics[width=\unitlength,page=16]{ContrasTR_inference.pdf}}%
    \put(0.81323783,0.21554048){\makebox(0,0)[lt]{\lineheight{1.25}\smash{\begin{tabular}[t]{l}N\end{tabular}}}}%
    \put(0,0){\includegraphics[width=\unitlength,page=17]{ContrasTR_inference.pdf}}%
  \end{picture}%
\endgroup%

%% file: experiments.tex
\begin{figure*}[t]
\setlength\tabcolsep{1pt} 
\centering
\begin{tabular}{@{} r M{0.31\linewidth} M{0.31\linewidth} M{0.31\linewidth} @{}}
    \begin{subfigure}{0.03\linewidth} \caption{}\label{subfig:bdd100k_vid1_c_main-paper} \end{subfigure} 
      & \includegraphics[width=\hsize]{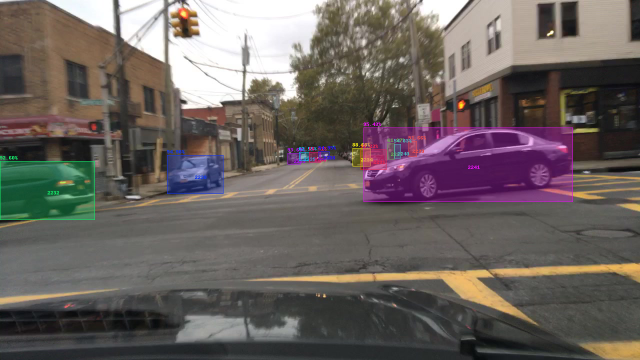} 
      & \includegraphics[width=\hsize]{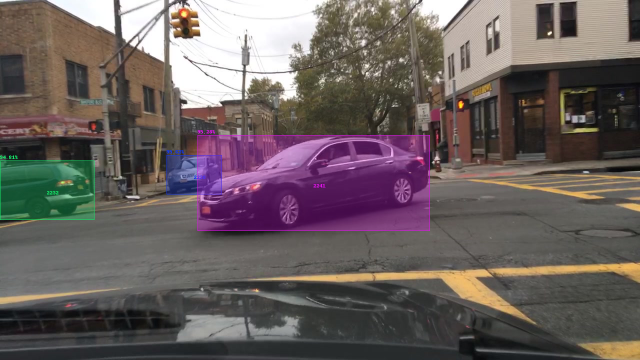}
      & \includegraphics[width=\hsize]{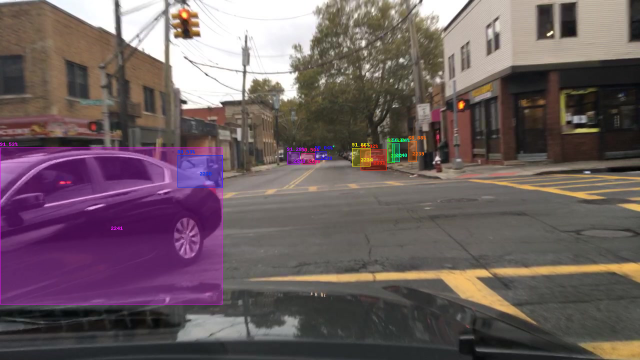}\\ \addlinespace
    \begin{subfigure}{0.03\linewidth} \caption{}\label{subfig:bdd100k_vid1_b_main-paper} \end{subfigure} 
      & \includegraphics[width=\hsize]{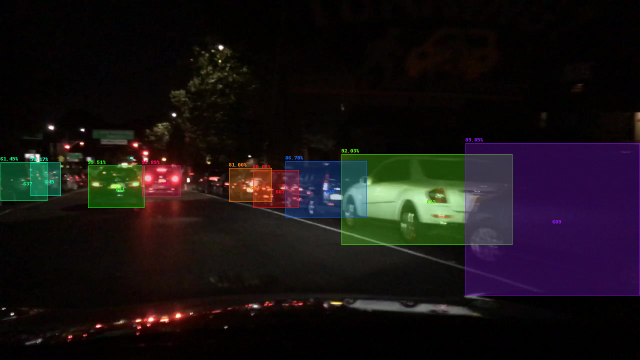} 
      & \includegraphics[width=\hsize]{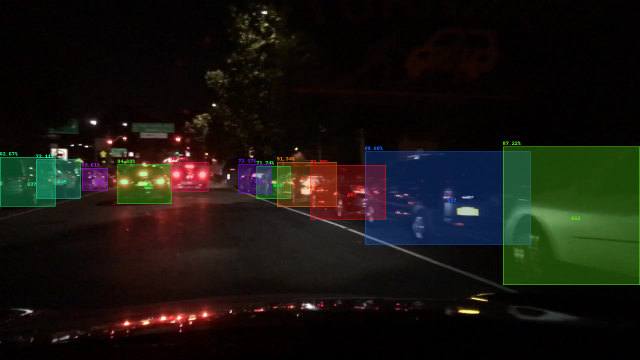}
      & \includegraphics[width=\hsize]{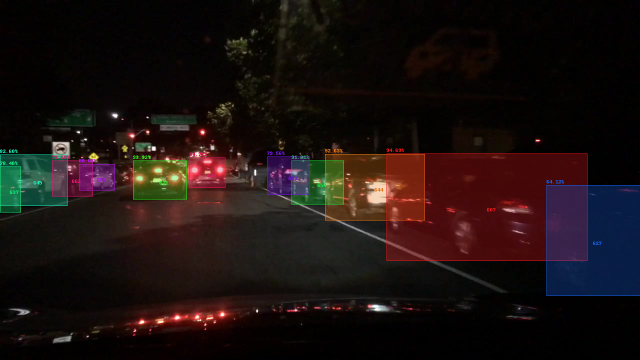}\\ 
\end{tabular}
   \caption{Predictions our model on the validation set of BDD100K, each color represents a different predicted ID. The method is robust to occlusions and in the nighttime.}
   \label{fig:bdd100k_qualitative}
   \vspace{-1.5em}
\end{figure*}

\section{Experiments}

In this section, we show the results of our method on the MOT17 \cite{MOT16} and BDD100K \cite{bdd100k} benchmarks. We then conduct an ablation study to obtain more detailed insights about our framework.

\subsection{Experimental setup}
\label{sec:experimental-setup}

\paragrax{Datasets.} MOT17 \cite{MOT16} is a widely used benchmark to evaluate \ac{mot} methods. It consists of training and test sets, each including seven sequences. Only pedestrians are evaluated in this benchmark.
\noindent BDD100K \cite{bdd100k} is a large-scale driving video dataset comprising 100K diverse video clips in different environmental and weather conditions. The dataset provides several types of annotations, such as object bounding boxes, semantic and instance segmentation, tracking IDs, etc. 
The object detection set consists of one annotated frame per video and the MOT set is a subset of 1600 videos, annotated at a lowered frame rate of 5 Hz.

\paragrax{Evaluation metrics.} The most widely used metric to evaluate \ac{mot} methods is \ac{mota} \cite{Bernardin2008EvaluatingMO}. However, the more recently introduced \ac{hota} \cite{Luiten2020IJCV} metric better balances detection and association. Since its introduction, many \ac{mot} benchmarks have adopted it, including MOT17 and BDD100K. \ac{idf1} \cite{10.1007/978-3-319-48881-3_2} is also widely used to measure \ac{mot} systems and it focuses on object's identities association. While on MOT17 the main metrics are computed over all the classes together, BDD100K computes these metrics independently for each class and then computes an average. As a result, the main metrics are defined as \ac{mhota}, \ac{mmota} and \ac{midf1}. 

\begin{table*}[t]
    \centering \footnotesize
    \begin{tabular}{llllllll}
         \toprule
        Method             & Backbone   & Pre-train & Train  & HOTA$\uparrow$ & MOTA$\uparrow$ & IDF1$\uparrow$  & IDS$\downarrow$ \\ 
        \midrule
        \textit{CNN-based}: &&&&&&& \\
        CenterTrack \cite{zhou2020tracking}    &  DLA-34 & CH      & MOT17      & 52.2   & 67.8      & 64.7      & 3039  \\
        QDTrack \cite{cvpr_qdtrack}    & ResNet-50 & COCO      & MOT17      & ---   & 68.7      & 66.3      & 3378  \\
        MTrack \cite{yu2022towards}    & DLA-34    & CH        & MOT17      & ---   & 72.1      & 73.5      & 2028  \\
        FairMOT \cite{zhang2021fairmot}& DLA-34    & COCO      & MOT17+CH+CP+ETHZ+CS+CT+PRW & 59.3  & 73.7      & 72.3      & 3303  \\ 
        Swin-JDE \cite{tsai2023swinJDE}& ResNet-50 & CH        & MOT17+CP+ETHZ+CS+CT+PRW & 57.2   & 71.7  & 71.3 & 2784 \\
        ByteTrack \cite{ByteTrack}     & YOLOX-X   & COCO      & MOT17+CH+CP+ETHZ  & 63.1 & 80.3   & 77.3      & 2196 \\
        SUSHI \cite{Cetintas_2023_CVPR}     & YOLOX-X   & COCO      & MOT17+CH+CP+ETHZ  & 66.5 & 81.1   & 83.1      & 1149 \\
         \midrule
        \textit{Transformer-based}: &&&&&&& \\
        MO3TR PIQ-SST \cite{zhu2022looking} & DarkNet & COCO      & MOT17+CH  & 60.5      & 78.6     & 72.4 & 2808 \\
        TransCenter \cite{9964258} &  PVTv2 & COCO & MOT17+CH & --- & 76.2 & 65.5 & 5394 \\
        MOTRv2  \cite{Zhang_2023_CVPR}     & YOLOX-X+ResNet-50 & COCO     & MOT17+CH*+CP+ETHZ   & 62.0      & 78.6      & 75.0      & 2619  \\
        \cdashlinelr{1-8}
        MO3TR \cite{zhu2022looking} & ResNet-50 & COCO      & MOT17+CH+ETHZ+CS  & 49.9      & 63.9     & 60.5          & 2847 \\
        MeMOT \cite{cai2022memot}          & ResNet-50 & COCO      & MOT17+CH*   & 56.9      & 72.5      & 69.0      & 2724  \\
        TrackFormer \cite{meinhardt2022trackformer} & ResNet-50 & CH        & MOT17+CH   & 57.3      & 74.1      & 68.0      & 2829  \\
        MOTR  \cite{zeng2022motr}          & ResNet-50 & COCO      & MOT17+CH*   & 57.8      & 73.4      & 68.6      & \textbf{2439}  \\
        TransTrack \cite{sun2020transtrack}& ResNet-50 & CH        & MOT17+CH   & --- & \textbf{74.5}      & 63.9      & 3663  \\
        TransTrack \cite{sun2020transtrack}& ResNet-50 & CH        & MOT17      & --- & 68.4      &  ---         & 3942  \\ 
        \textbf{\ourmodelname{}} (ours)  & ResNet-50 & CH        & MOT17      & \textbf{58.9}     & 73.7      & \textbf{71.8}     & 2619  \\
         \bottomrule
    \end{tabular}
    \caption{Results on the MOT17 test set using private detections. The second group shows DETR-like models, all based on Deformable-DETR except MO3TR, which is based on DETR. The best results among DETR-like models using the ResNet-50 backbone are highlighted in \textbf{bold}. Datasets abbreviations refer to the following works  ETHZ \cite{4587581}, CityPersons \cite{8099957}, CS \cite{8099843}, CT \cite{dollar2009caltech_dataset}, PRW \cite{zheng2017PRW_dataset}. 
    }
    \label{tab:metrics-on-MOT17}
\end{table*}

\vspace{-0.5em}

\subsection{Implementation details} 

Our method is built on top of Deformable-DETR \cite{zhu2020deformable} with a ResNet-50 \cite{7780459} backbone. We also experiment with a more powerful Swin-L \cite{liu2021swin} backbone to demonstrate the scalability of our method.
The architecture includes \textit{mixed selection} and \textit{look forward twice} refinements introduced in DINO \cite{zhang2022dino}. Unless stated otherwise, we train the models with their default hyper-parameter sets. More details are provided in Appendix \ref{app:hyper_params}. 
Our additional loss term is computed over every transformer decoder layer output, and the temperature value $\tau$ is set to $0.1$. 
The \textit{objectness threshold} is set to 0.5 during inference.

Our model is pre-trained on the object detection task with the procedure described in Section \ref{sec:learnmotfromdet} with the default Deformable-DETR hyperparameters and augmentations. The contrastive loss coefficient is set to $\lambda_{\mathrm{contr}} = 2$. We did not use the contrastive pre-training when pre-trained on COCO \cite{lin2014microsoft}.

\paragrax{MOT17.} We follow a similar procedure described in \cite{zhou2020tracking} by pre-training the model on CrowdHuman (CH)\cite{shao2018crowdhuman}. However, we do not simulate tracking frames. For simplicity of the training procedure, we skip the joint MOT17+CH dataset fine-tuning stage and only use MOT17. 
We pre-train the model for 50 epochs with a batch size of $8$ and augment each image twice for the contrastive loss, leading to a total batch size of $16$. In the training stage, the batches are built by sampling $N_v=2$ videos and $N_f=8$ frames, and $\lambda_{\mathrm{contr}}$ is set to 2. In addition, we allow past detected objects to be kept in memory for a maximum of $T=20$ frames, and we evaluate our method under the private detection protocol.

\paragrax{BDD100K.} Our model is pre-trained on the BDD100K detection dataset for 36 epochs. We use a batch size of 24 and again augment each image twice, leading to a total batch size of 48. 
Afterwards, we train it on the tracking set for 10 epochs and decrease the learning rate by a factor of 10 at epoch 8. Each training batch samples $N_v=4$ videos and $N_f=10$ frames. The contrastive loss coefficient is set to $\lambda_{\mathrm{contr}} = 1$. In addition, we allow past detected objects to be kept in memory for a maximum of $T=9$ frames.

\vspace{-0.5em}

\begin{table*}[t]
\centering \footnotesize
\begin{tabular}{lllllll}
\toprule
Method                              & Backbone  & Pre-train & mHOTA$\uparrow$ & mMOTA$\uparrow$ & mIDF1$\uparrow$ & IDS$\downarrow$ \\ \midrule
\textit{Yu et al.} \cite{bdd100k}   & ResNet-101 & ---        & ---    & 26.3     & 44.7      & 14674 \\ 
QDTrack \cite{cvpr_qdtrack}              & ResNet-50 & BDD100K   & 41.8   & 35.6     & 52.4      & \textbf{10790} \\ 
TETer \cite{li2022TETer_tracking}   & ResNet-50 & BDD100K   & ---    & 37.4     & 53.3      & --- \\
ByteTrack \cite{ByteTrack}          & YOLOX-X   & COCO      & ---    & 40.1     & 55.8      & 15466 \\
SUSHI \cite{Cetintas_2023_CVPR}     & YOLOX-X   & COCO      & \textbf{48.2}    & 40.2     & \textbf{60.0}      & 13626 \\
\textbf{\ourmodelname{}}  (ours)              & Swin-L    & BDD100K   & 46.1 & \textbf{42.8} & 56.5     & 10793   \\
\bottomrule
\end{tabular}
\caption{Results on BDD100K test split, with an objectness threshold of 0.4. The best results are shown in \textbf{bold}.}
\label{tab:metrics-on-BDD100K-test}
\vspace{-2.0em}
\end{table*}

\begin{table*}[t]
\centering \footnotesize
\begin{tabular}{lllllll}
\toprule
Method                                       & Backbone  & Pre-train & mHOTA$\uparrow$ & mMOTA$\uparrow$ & mIDF1$\uparrow$ & IDS$\downarrow$ \\ \midrule               
MOTR \cite{zeng2022motr}        & ResNet-50 & COCO      & ---             & 32.0            & 43.5            & \textbf{3493}   \\
MOTRv2 \cite{Zhang_2023_CVPR}   & ResNet-50 & COCO      & ---             & 35.5            & 48.2            & ---  \\
\textbf{\ourmodelname{}} w/o CPT (ours)  & ResNet-50 & COCO      & 40.2            & 36.4            & 48.1            & 6067  \\ 
\textbf{\ourmodelname{}} (ours)          & ResNet-50 & BDD100K & \textbf{40.8} & \textbf{36.7} & \textbf{49.2} &  6695 \\ \midrule 
MOTRv2 \cite{Zhang_2023_CVPR}   & YOLOX-X+ResNet-50 & BDD100K+COCO & ---       & \textbf{43.6}     & \textbf{56.5}            & ---  \\ 
\textbf{\ourmodelname{}} (ours)          & Swin-L & BDD100K   & 44.4   & 41.7   & 52.9   & 6363            \\ 
\bottomrule
\end{tabular}
\caption{Results on the BDD100K validation set, with an objectness threshold of 0.4. The best results are highlighted in \textbf{bold}. } 
\label{tab:metrics-on-BDD100K-val}
\vspace{-1.4em}
\end{table*}

\subsection{Results}

\paragrax{MOT17.} We report quantitative results on the test set of MOT17 in Table \ref{tab:metrics-on-MOT17}. Our method outperforms comparable DETR-like models (bottom section) on \ac{hota}, suggesting that our model has the best balance between detection and association. It also outperforms comparable DETR-like models on \ac{idf1} which shows the effectiveness of the object representations learned through the instance-level contrastive loss. 

\paragrax{BDD100K.} For BDD100K, we report results on the test set in Table \ref{tab:metrics-on-BDD100K-test}. 
Our method outperforms all methods by a significant margin on the mMOTA metric, including unpublished methods\footnote{The full leaderboard, including unpublished methods, is available on \textit{eval.ai} \cite{evalAI:Online}.}.
We further report results on the validation set (Table \ref{tab:metrics-on-BDD100K-val}) to compare with MOTR \cite{zeng2022motr} and MOTRv2 \cite{Zhang_2023_CVPR}, since the methods were not evaluated on the test set.
MOTRv2 outperforms our method when using an additional YOLOX \cite{ge2021yolox} detector for object proposals, which highly reduces the number of misses.
Nevertheless, our model with a ResNet-50 backbone outperforms MOTR \cite{zeng2022motr} and MORTv2 \cite{Zhang_2023_CVPR}. 
To make a fair comparison, we also pre-train our model on COCO. Although we don't use contrastive pre-training in that setting, we still exceed MOTRv2's performance by +0.3 \ac{mmota} and +1.1 \ac{midf1}. 

\paragrax{Limitations.} Yet, our method does not achieve the same level of performance as the state-of-the-art tracking methods on MOT17. As shown in the table, differences can be attributed to extra training data, more powerful backbones, extra post-processing and more elaborate association strategies. 

\vspace{-0.2em}

\subsection{Ablation study}

We conduct the ablation study on BDD100K \cite{bdd100k} as the dataset is large and offers adequate variation. We pre-train our models on the detection set of BDD100K without the contrastive loss unless specified, and then we train on $1/8$ of the tracking training data by randomly subsampling $25\%$ of the videos and $50\%$ of the frames. The entire validation set is used for evaluation. We use a batch size of 40, $T=5$ previous frames and an \textit{objectness threshold} of 0.5. Due to limited computational resources, the ablations have not been run with the optimal parameters.

\paragrax{Sampling strategy.} In this ablation, we investigate different variations of the number of sampled videos $N_v$ and sampled frames $N_f$ to show the effect of increasing one variable and simultaneously decreasing the other. For the experiment, we used a batch size of $N_vN_f=40$ and a contrastive loss weighting of $\lambda_{\mathrm{contr}} = 2$. Results are given in Table \ref{tab:samples-variation}. As the number of sampled videos $N_v$ decreases and the number of sampled frames per video $N_f$ increases, the probability of sampling larger sets of positive examples rises. This explains the increasing tendency of tracking metrics that emphasize association: the contrastive loss can leverage a greater variety of positive and negative examples. 
However, when sampling all 40 frames within one video, the scenery variability decreases and so the variation in the set of negative examples dramatically drops, harming performance.

\begin{table}[h!]
\centering \footnotesize
\begin{tabular}{lllllll}
\toprule
$N_v$ & $N_f$ & mHOTA & mMOTA & mIDF1 & mAP & IDSw \\
\midrule
20 & 2  & 35.0     & 33.3     & 40.8     & 33.8     & 10116     \\
8  & 5  & 35.6     & 33.3     & 41.6     & 33.8     & 8724     \\
4  & 10 & 35.9     & \textbf{33.7}     & 42.3     & 33.7     & 7869     \\
2  & 20 & \textbf{36.2}     & 33.6     & \textbf{42.9}     & 33.7     & 7483     \\
1  & 40 & 35.4     & 32.1     & 41.3     & \textbf{34.0}     & \textbf{7191}     \\
\bottomrule
\end{tabular}
\caption{Influence of the number of frames sampled per video on BDD100K validation set. The batch size $N_v N_f$ is fixed at 40.}
\label{tab:samples-variation}
\vspace{-1em}
\end{table}

\paragrax{Contrastive loss weight.}
The contrastive loss parameter $\lambda_{\mathrm{contr}}$ also impacts performance. We experimented with different values and found values close to 0.5 to offer the best results. When the coefficient is too high, detection performance suffers, so overall tracking performance decreases.
In practice, we found a slight shift in the optimal value when using contrastive pre-training and the whole training data.

\vspace{-0.5em}
\begin{table}[h!]
\centering \footnotesize
\begin{tabular}{llllll}
\toprule
$\lambda_{\mathrm{contr}}$ & mHOTA & mMOTA & mIDF1 & mAP & IDSw \\ \midrule
0.25 & 36.4         & 34.9          & 42.9          & \textbf{34.8}      & 10105      \\
0.5 & \textbf{36.6}      & \textbf{35.0}      & \textbf{43.1}     & 34.7     & 9216 \\
1   & 36.5      & 34.5      & \textbf{43.1}     & 34.4     & 8502 \\
2   & 35.9      & 33.7      & 42.3     & 33.7     & 7869    \\
3   & 35.0      & 32.2      & 40.9     & 33.2     & 7693 \\
4   & 34.3      & 31.0      & 39.8     & 32.7     & \textbf{7550} \\
\bottomrule
\end{tabular}
\caption{Influence of the contrastive loss weighting on object detection and MOT metrics on BDD100K validation set. Classification and localization coefficients are kept fixed. A higher coefficient reduces the number of identity switches but comes at the cost of a lower mAP. All models have been trained with $N_f=10$, $N_v=4$.} 
\label{tab:contr-loss-weight-variation}
\vspace{-1em}
\end{table}

\paragrax{Contrastive learning, pre-training and projection head.}
We ablate the effect of our contributions on the tracking metrics in Table \ref{tab:contr-proj-cpt}. All models are trained with $N_f=10$, $N_v=4$ and $\lambda_{\mathrm{contr}}=1$.
Without the contrastive loss, the model is not trained for tracking and produces many ID switches (fourth row).
When the model is trained without the FFN projection head, the contrastive loss is applied directly over the embeddings produced by the transformer decoder. In this setup, the embeddings produced by the transformer decoder are used to generate the bounding boxes and class predictions through the respective heads and as tracking embeddings to re-identify objects over time. The performance gap (between the second and third row) shows that this additional head provides more flexibility to the network to generate the tracking embeddings and preserves detection performance.
Pre-training the model with the contrastive loss on object detection using the methodology outlined in Section \ref{sec:learnmotfromdet} improves tracking metrics significantly (first row).
Finally, the performance difference between Deformable-DETR (last row) and ContrasTR (first row) is small in terms of mMOTA. This is because mMOTA overemphasizes the effect of detection performance \cite{Luiten2020IJCV}.

\vspace{-0.5em}

\begin{table}[h!]
\centering \footnotesize
\begin{tabular}{cccllll}
\toprule
Contr. & Proj. & CPT & mHOTA & mMOTA & mIDF1 & mAP \\ \midrule
\cmark  & \cmark  & \cmark    & \textbf{38.0}     & \textbf{35.1} & \textbf{45.1} & 34.3 \\ 
\cmark  & \cmark  & \xmark    & 36.5     & 34.5   & 43.1    & \textbf{34.4} \\
\cmark  & \xmark  & \xmark    & 35.4     & 33.8   & 41.4    & 33.6 \\
\xmark  & \xmark  & \xmark    & 33.4     & 32.9   & 38.3    & 34.3 \\
\bottomrule
\end{tabular}
\caption{Ablation study of different method components on BDD100K validation set: the contrastive loss and the sampling strategy (Contr.); the tracking projection head (Proj.); the contrastive pre-training on object detection (CPT).}
\label{tab:contr-proj-cpt}
\vspace{-1.5em}
\end{table}

%% file: conclusion.tex
\vspace{-0.5em}

\section{Conclusion}
This work presents a joint object detection and \ac{mot} method, \ourmodelname{}. Our model exploits DETR's object-level embeddings to learn robust object representations through an instance-level contrastive loss and a revised sampling strategy.
The proposed method preserves detection performance and can turn existing DETR-like object detectors into trackers without needing complicated architectural design or additional heavy modules. We also present a highly scalable pre-training scheme to exploit detection-only datasets to boost performance further.

It matches the performance of comparable DETR-like methods on the MOT17 benchmark while using less additional training data.
Furthermore, its performance surpasses the previous state-of-the-art by +2.6 mMOTA on the challenging BDD100K dataset. Thereby demonstrating its robustness in adverse weather conditions and under significant object motion.

\paragraph{Acknowledgements} The authors thankfully acknowledge support by Toyota via the TRACE project. All the authors are also affiliated to Leuven.AI - KU Leuven institute for AI, B-3000, Leuven, Belgium.

%% file: supplementary.tex
\begin{center}
{\LARGE \textbf{Contrastive Learning for Multi-Object Tracking with Transformers}}
\vspace{7pt}
\\ {\large \textbf{Supplementary Material}}
\end{center}

\section{Influence of the memory length}

Tracking embeddings are kept in the memory for a given number of previous frames $T$, and they allow object re-identification without the need for learnable parameters to update tracks. The influence of the maximum memory length on the tracking metrics for BDD100K and MOT17 is displayed in Table \ref{tab:memory-length_BDD100K} and Table \ref{tab:memory-length_MOT17}. A history of one previous frame leads to low results on both datasets. The optimal memory length varies from around 9 previous frames on BDD100K to around 30 previous frames on MOT17. Since the parameter $T$ only influences the association task, it has to be sufficiently high to recover IDs after occlusions or miss detections but not too high to avoid picking up an ID of an object that has left the scene. We state the difference between BDD100K and MOT17 mostly comes from the frame rates, which are respectively 5Hz and 30 Hz\footnote{On MOT17, videos 1-4 and 7-12 have a frame rate of 30 Hz, whereas videos 5-6 and 13-14 have a frame rate of respectively 14Hz and 25 Hz.}. In seconds, the optimal memory durations are respectively 1.8s and 1s. 

\begin{table}[h!]
\centering \footnotesize
\begin{tabular}{lllll}
\toprule
length & mHOTA$\uparrow$ & mMOTA$\uparrow$ & mIDF1$\uparrow$ & IDSw$\downarrow$ \\ \midrule
1   & 34.6      & 33.1      & 38.8     & 15404      \\
5   & 38.0      & 35.1      & 45.1     & 6827       \\ 
9   & \textbf{38.4}      & 35.3      & \textbf{45.8}     & 6186       \\
12  & 38.3          & \textbf{35.4}          & 45.4         & \textbf{6092}           \\
15  & 38.2          & \textbf{35.4}          & 45.2         &  6153           \\
\bottomrule
\end{tabular}
\caption{Influence of the memory length on MOT metrics on BDD100K validation set.}
\label{tab:memory-length_BDD100K}
\end{table}

\begin{table}[h!]
\centering \footnotesize
\begin{tabular}{lllll}
\toprule
length & HOTA$\uparrow$ & MOTA$\uparrow$ & IDF1$\uparrow$ & IDSw$\downarrow$ \\ \midrule
1   & 57.0      & 72.2      & 63.6     & 1102      \\
10   & 62.3      & \textbf{73.6}      & 73.7     & 368      \\
20   & 63.0      & \textbf{73.6}      & 75.7     & 340      \\
30   & \textbf{63.5}      & \textbf{73.6}      & \textbf{76.4}     & \textbf{331}      \\
40   & 63.2      & \textbf{73.6}      & 76.2     & 343      \\
\bottomrule
\end{tabular}
\caption{Influence of the memory length on MOT metrics on MOT17 validation set.}
\label{tab:memory-length_MOT17}
\end{table}

\begin{figure*}[t]
\setlength\tabcolsep{1pt}
\centering
\begin{tabular}{@{} r M{0.31\linewidth} M{0.31\linewidth} M{0.31\linewidth} @{}}
    \begin{subfigure}{0.05\linewidth} \caption{}\label{subfig:bdd100k_vid1_a2} \end{subfigure} 
      & \includegraphics[width=\hsize]{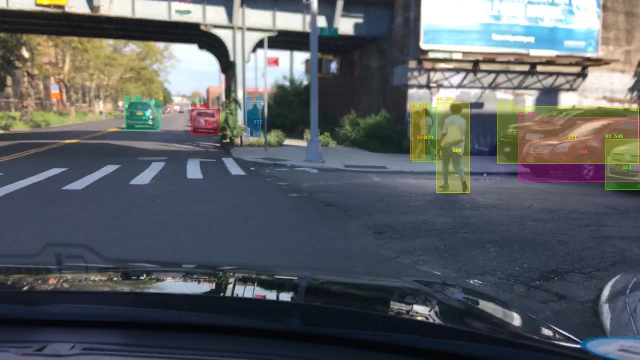} 
      & \includegraphics[width=\hsize]{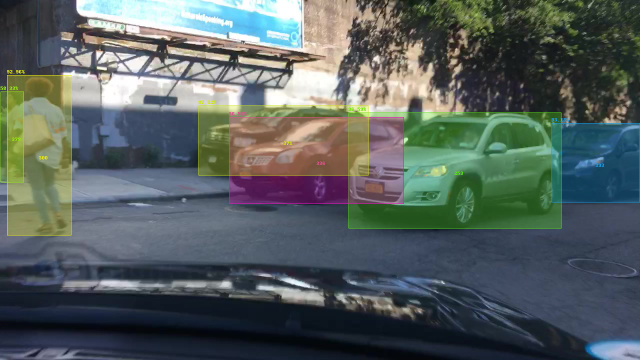}
      & \includegraphics[width=\hsize]{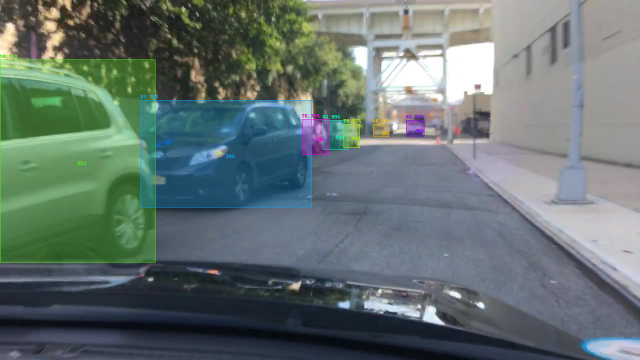}\\ \addlinespace
    \begin{subfigure}{0.05\linewidth} \caption{}\label{subfig:bdd100k_vid1_b} \end{subfigure} 
      & \includegraphics[width=\hsize]{16861.png} 
      & \includegraphics[width=\hsize]{16865.png}
      & \includegraphics[width=\hsize]{16869.png}\\ \addlinespace
    \begin{subfigure}{0.05\linewidth} \caption{}\label{subfig:bdd100k_vid1_c} \end{subfigure} 
      & \includegraphics[width=\hsize]{17911.png} 
      & \includegraphics[width=\hsize]{17915.png}
      & \includegraphics[width=\hsize]{17920.png}\\ \addlinespace
    \begin{subfigure}{0.05\linewidth} \caption{}\label{subfig:bdd100k_vid1_d} \end{subfigure} 
      & \includegraphics[width=\hsize]{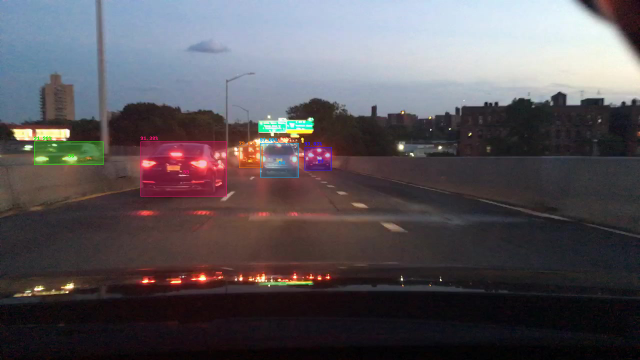} 
      & \includegraphics[width=\hsize]{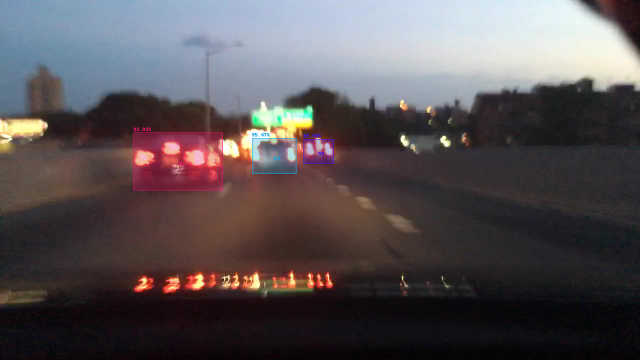}
      & \includegraphics[width=\hsize]{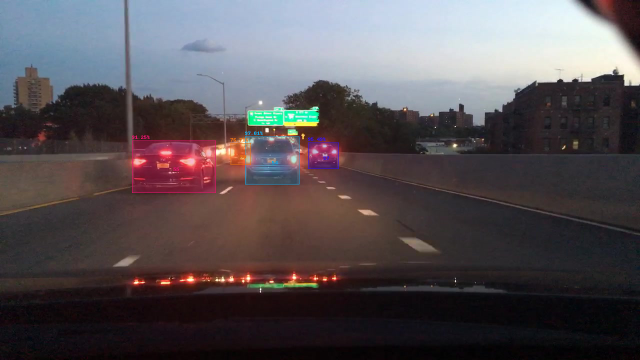}\\ \addlinespace
    \begin{subfigure}{0.05\linewidth} \caption{}\label{subfig:bdd100k_vid1_g} \end{subfigure} 
      & \includegraphics[width=\hsize]{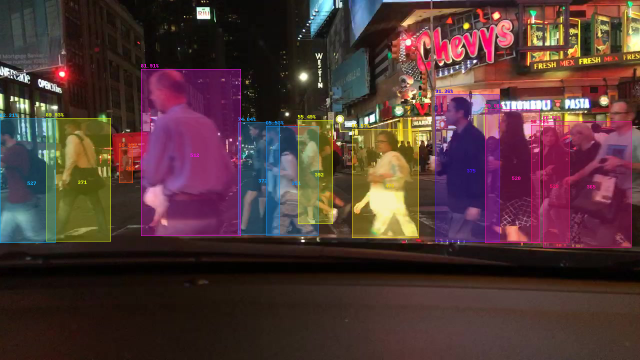} 
      & \includegraphics[width=\hsize]{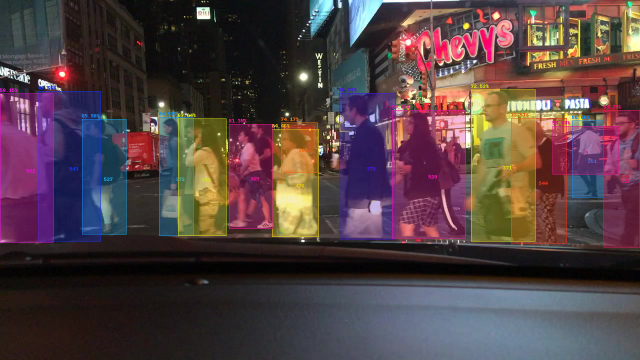}
      & \includegraphics[width=\hsize]{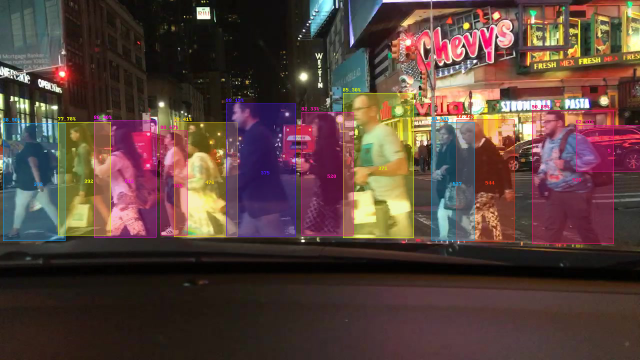}\\ \addlinespace
    \begin{subfigure}{0.05\linewidth} \caption{}\label{subfig:bdd100k_vid1_h} \end{subfigure} 
      & \includegraphics[width=\hsize]{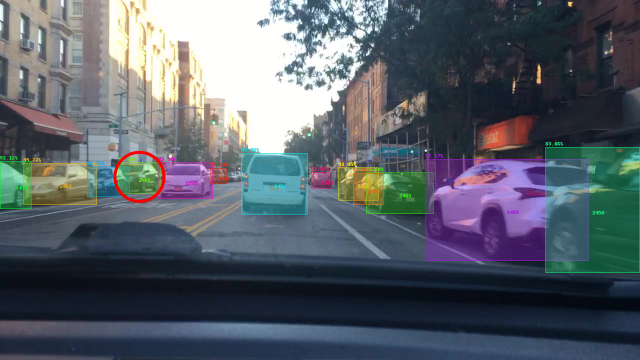} 
      & \includegraphics[width=\hsize]{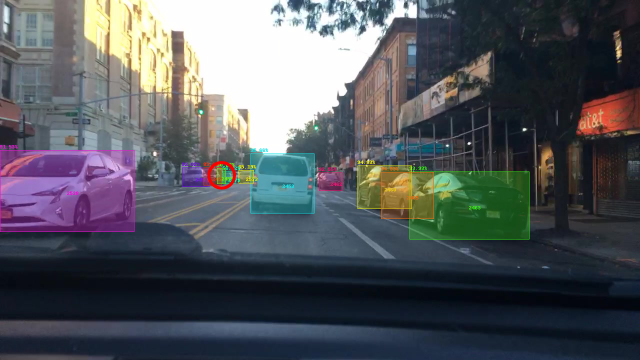}
      & \includegraphics[width=\hsize]{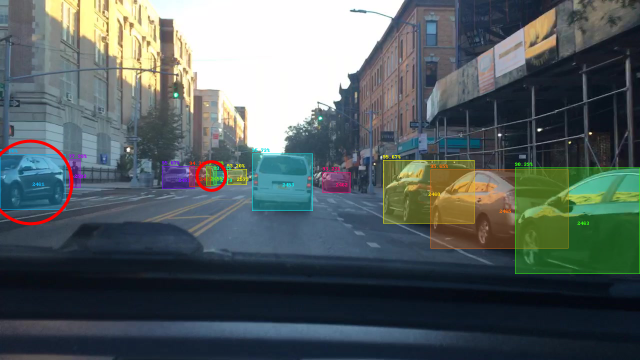}\\ \addlinespace
\end{tabular}
   \caption{Predictions and failure cases of our model on the validation set of BDD100K.}
   \label{fig:bdd100k_vid1}
\end{figure*}

\section{Qualitative results}

Figure \ref{fig:bdd100k_vid1} shows predictions from \ourmodelname{} on the validation set of BDD100K. 
Since our method learns instance-level features by exploiting different views of the same object, it is robust in case of large camera movement (Figure \ref{subfig:bdd100k_vid1_a2}). It also performs well in case of (partial) object occlusion (Figure \ref{subfig:bdd100k_vid1_c}). 
Furthermore, the method predicts discriminative tracking embeddings even under night conditions (Figure \ref{subfig:bdd100k_vid1_b}) and in case of motion blur (Figure \ref{subfig:bdd100k_vid1_d}).
Nevertheless, the method sporadically swaps or re-assigns IDs from disappeared pedestrians in heavily crowded scenes (Figure \ref{subfig:bdd100k_vid1_g}), or it assigns an ID from an occluded object to a newly appeared one. In Figure \ref{subfig:bdd100k_vid1_h}, the car highlighted with a red circle (first frame) is occluded in the second frame, and its ID is assigned to a car further away. After the occlusion, the car further away kept its ID, and the first car got a new ID. Incorporating into the model trajectory and temporal information could be beneficial in mitigating these failure cases. 

\begin{figure*}[t]
\setlength\tabcolsep{5pt} 
\centering
\begin{tabular}{@{} r M{0.60\linewidth} M{0.31\linewidth} @{}} 
    \begin{subfigure}{0.05\linewidth} \caption{}\label{subfig:bdd100k_embeddings_video-170} \end{subfigure} 
      & \includegraphics[width=\hsize]{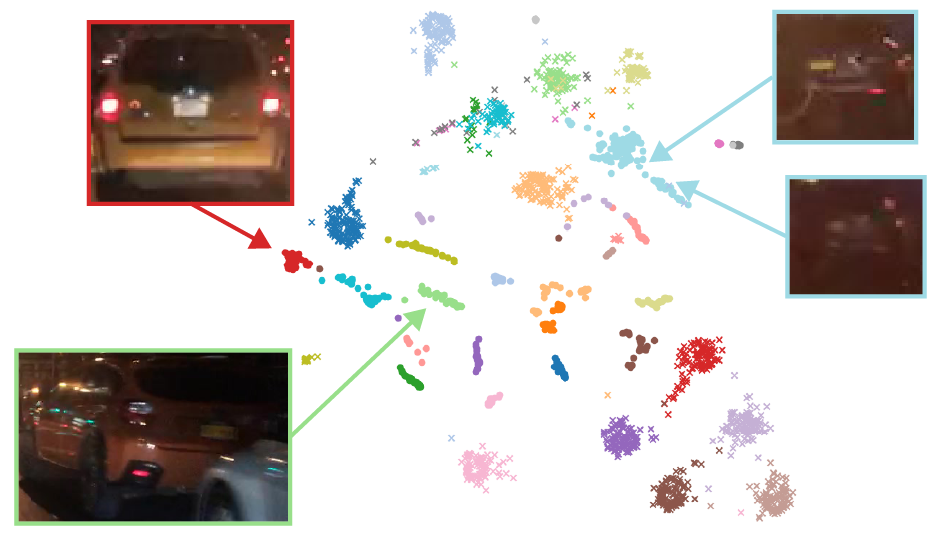}
      & \includegraphics[width=\hsize]{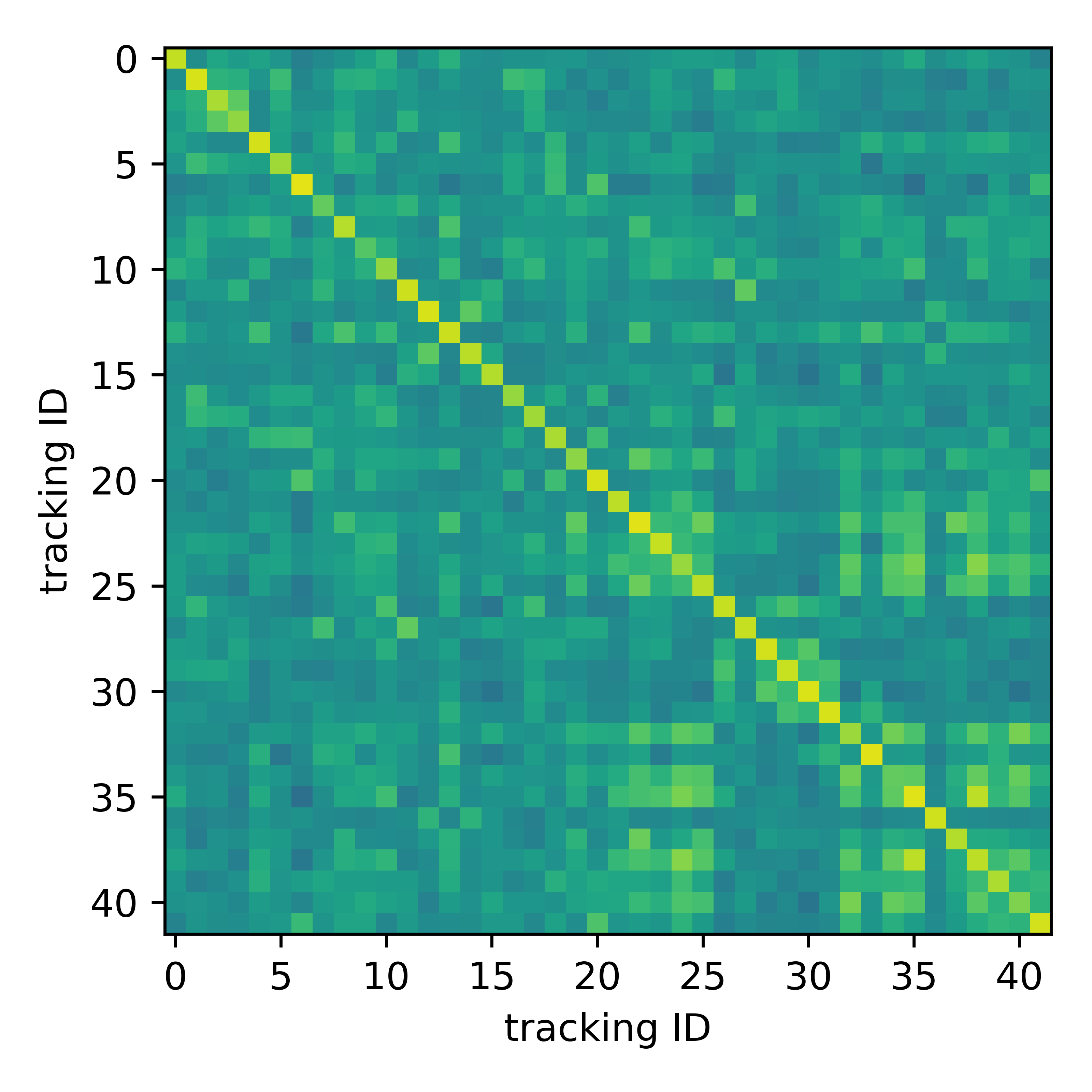}
      \\ \addlinespace
    \begin{subfigure}{0.05\linewidth} \caption{}\label{subfig:bdd100k_embeddings_video-166} \end{subfigure}
      & \includegraphics[width=\hsize]{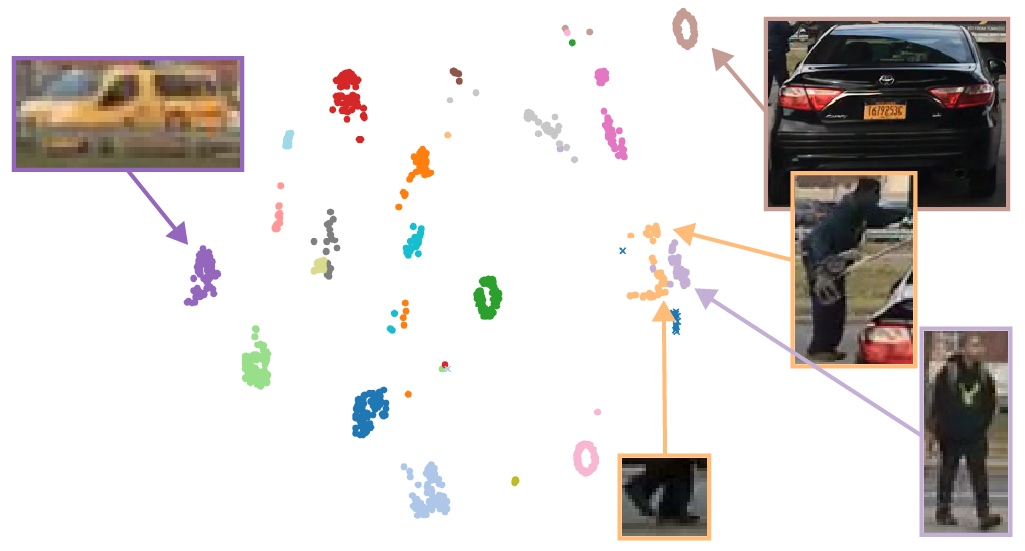}
      & \includegraphics[width=\hsize]{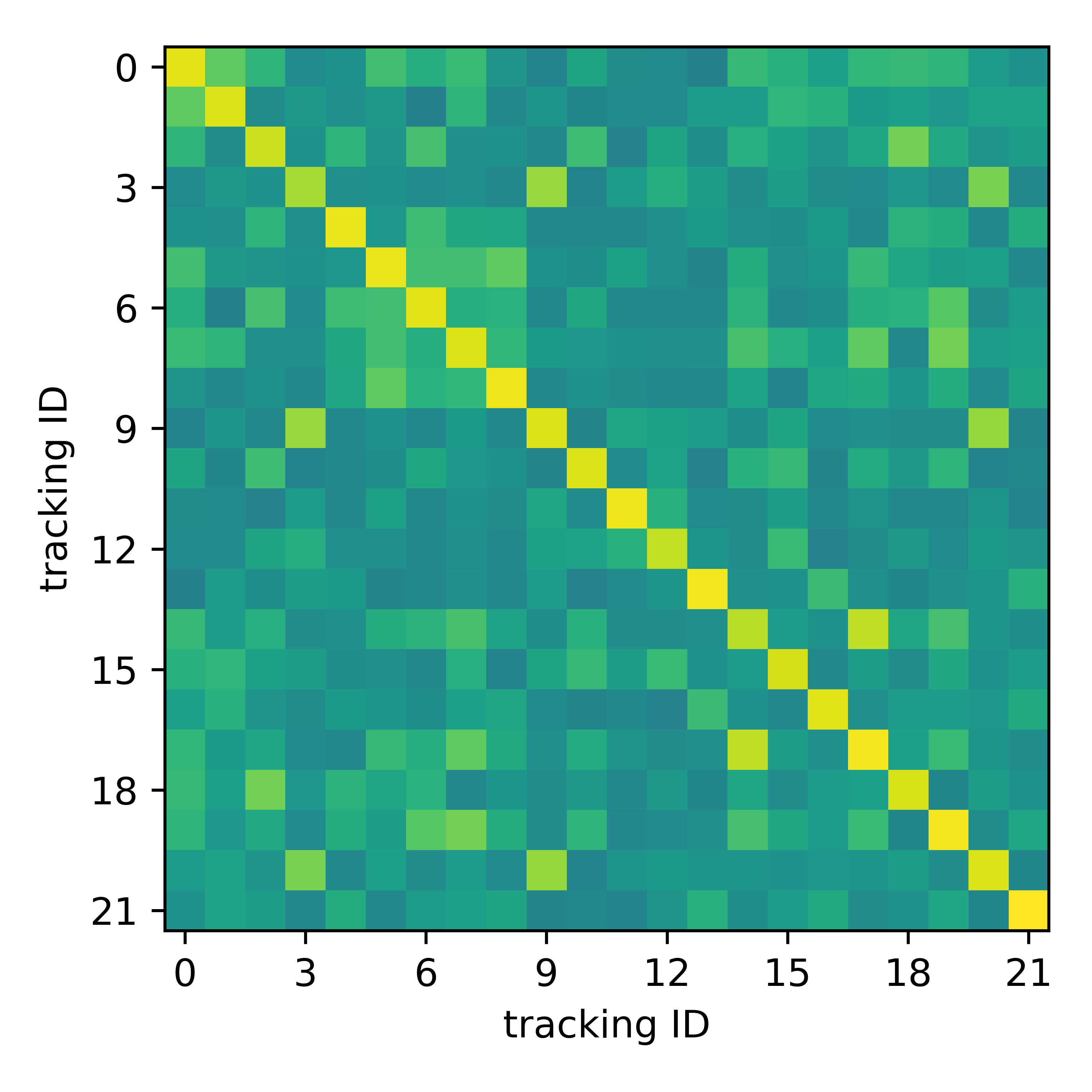}
      \\ \addlinespace
    \begin{subfigure}{0.05\linewidth} \caption{}\label{subfig:bdd100k_embeddings_video-152} \end{subfigure}
      & \includegraphics[width=\hsize]{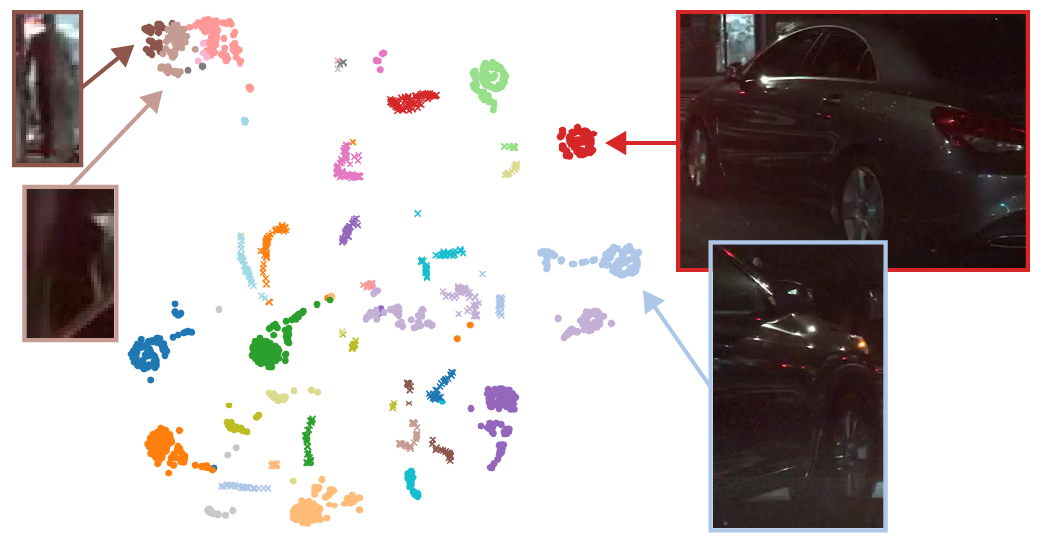} 
      & \includegraphics[width=\hsize]{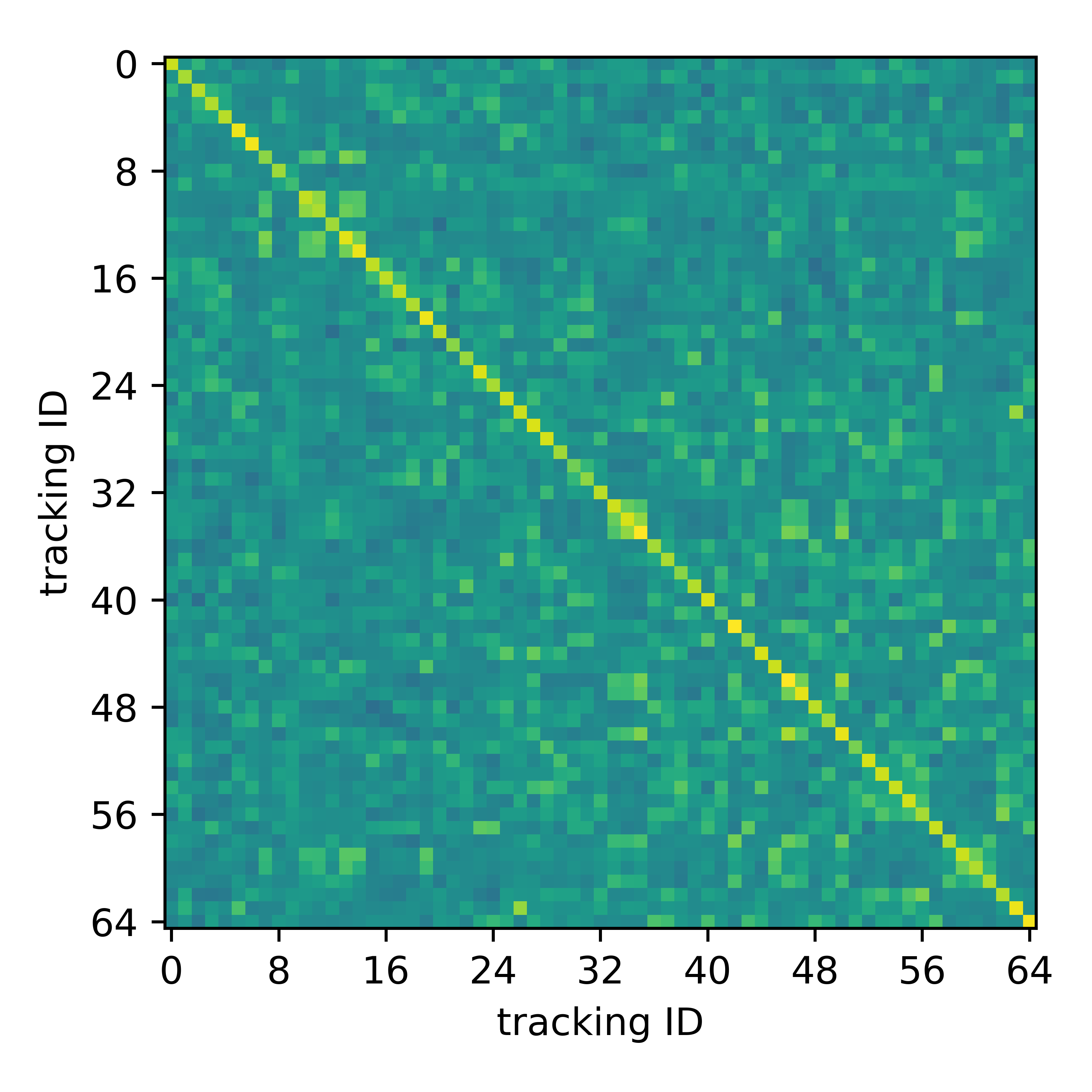} 
      \\ \addlinespace
\end{tabular}
   \caption{Visualization of predicted tracking embeddings on 3 different sequences from the BDD100K validation set. On the left, a t-SNE projection of the embeddings for the first 40 ground-truth objects, where each color-symbol pair represents a ground-truth tracking ID assigned with DETR’s bipartite matching. On the right, the average cosine similarity of predicted objects associated with the same ground-truth instance ID.}
   \label{fig:bdd100k_embeddings}
\end{figure*}

\section{Additional tracking embeddings visualization}

Figure \ref{fig:bdd100k_embeddings} shows a t-SNE projection (left) and the average cosine similarity (right) for the predicted tracking embeddings on three different videos from BDD100K.
Tracking IDs are assigned with DETR’s bipartite matching. Therefore, every ground truth object will be assigned to the tracking embedding with minimum object detection matching cost.
One can see that most tracking embeddings are clustered per ground-truth ID, even during night conditions (Figure \ref{subfig:bdd100k_embeddings_video-170} and Figure \ref{subfig:bdd100k_embeddings_video-152}).

\section{Hyper-parameters}
\label{app:hyper_params}
We report the hyper-parameters used for the experiments on MOT17 and BDD100K in Table \ref{tab:all_hyperparams}.

\begin{table*}
\centering \footnotesize
\begin{tabular}{lrrr}
\toprule
                             & MOT17           & BDD100K          & BDD100K    \\ \midrule
\textit{model}               & \multicolumn{3}{c}{Deformable-DETR with all refinements from DINO} \\
\textit{backbone}            & ResNet-50       & ResNet-50        & Swin-L     \\
\textit{classification head} & linear layer    & linear layer     & linear layer \\
\textit{localization head}   & 3 layers MLP    & 3 layers MLP     & 3 layers MLP  \\
\textit{tracking head}       & 3 layers MLP    & 3 layers MLP     & 3 layers MLP  \\
\textit{hidden dimension}    & 256             & 256              & 256        \\
\textit{dim feedforward (heads)}        & 256             & 256              & 256        \\
\textit{dim feedforward (transformer)}  & 1024            & 1024             & 1024       \\
\textit{num heads}           & 8               & 8                & 8          \\
\textit{num queries}         & 300             & 300              & 300        \\
\textit{weight decay}        & 0.05            & 0.05             & 0.05       \\
\textit{dropout}             & 0               & 0                & 0          \\
\textit{clip max norm}       & 0.1             & 0.1              & 0.1        \\
\textit{cls loss coef}       & 2               & 2                & 2          \\
\textit{bbox loss coef}      & 5               & 5                & 5          \\
\textit{giou loss coef}      & 2               & 2                & 2          \\
\textit{focal alpha}         & 0.25            & 0.25             & 0.25       \\
\textit{contrastive loss temperature} $\tau$ & 0.1    & 0.1           & 0.1    \\
\midrule \multicolumn{4}{l}{Pre-training} \\
\textit{dataset}             & CrowdHuman      & BDD100k detection & BDD100k detection    \\
\textit{epochs}              & 50              & 36               & 36       \\
\textit{LR drop (epoch)}     & None            & None             & None     \\
\textit{LR}                  & $2 \cdot 10^{-4}$ & $2 \cdot 10^{-4}$ & $2 \cdot 10^{-4}$  \\
\textit{LR backbone}         & $2 \cdot 10^{-5}$ & $2 \cdot 10^{-5}$ & $2 \cdot 10^{-5}$  \\
\textit{LR linear proj mult} & 0.1             & 0.1              & 0.1      \\
\textit{batch size}          & 16              & 48               & 24       \\
\textit{contrastive loss coef} & 2             & 2                & 2         \\
\midrule \multicolumn{4}{l}{Training} \\
\textit{epochs}              & 15              & 10               & 10       \\
\textit{LR drop (epoch)}     & 10              & 8                & 8        \\
\textit{LR}                  & $2 \cdot 10^{-5}$ & $2 \cdot 10^{-5}$ & $2 \cdot 10^{-5}$ \\
\textit{LR backbone}         & $2 \cdot 10^{-6}$ & $2 \cdot 10^{-6}$ & $2 \cdot 10^{-6}$ \\
\textit{LR linear proj mult} & 0.1             & 0.1              & 0.1      \\
\textit{batch size}          & 16              & 40               & 32       \\
\textit{contrastive loss coef} & 2             & 1                & 1        \\
\textit{num frames per video} $N_f$ & 8        & 10               & 8        \\
\textit{objectness threshold}& 0.5             & 0.4              & 0.4      \\
\textit{memory length} $T$     & 20            & 9                & 9        \\
\textit{new instance id threshold}& 0.5        & 0.5              & 0.5      \\

\bottomrule
\end{tabular}
\caption{Full set of hyper-parameters.}
\label{tab:all_hyperparams}
\end{table*}

\section{Additional details on benchmarks results}

\paragraph{MOT17.} We report in Table \ref{tab:metrics-on-MOT17-val-extended} the results of our method obtained on the validation set. The validation set has been selected following \cite{zhou2020tracking} and the training setup follows the one presented in Section \ref{sec:experimental-setup}. We also include a detailed Table \ref{tab:metrics-on-MOT17-test-extended} that includes more sub-metrics and video-level performance on the test set of MOT17.

\paragraph{BDD100K.} We also report detailed tables that include sub-metrics on the validation set and test set of BDD100K in Table \ref{tab:metrics-on-BDD100K-val-extendend} and in Table \ref{tab:metrics-on-BDD100K-test-extendend} respectively. The training setup follows the one presented in Section \ref{sec:experimental-setup}.

\begin{table*}[]
\centering \footnotesize
\begin{tabular}{llllllllllll}
\toprule
Sequence & HOTA$\uparrow$ & MOTA$\uparrow$ & IDF1$\uparrow$ & MT$\uparrow$  & ML$\downarrow$  & FP$\downarrow$    & FN$\downarrow$     & Rcll$\uparrow$ & Prcn$\uparrow$ & ID Sw.$\downarrow$  & Frag$\downarrow$   \\ \midrule
MOT17-02 & 45.0 & 50.1 & 54.6 & 12  & 12 & 413     & 4421    & 55.3    & 93.0    & 91  & 254  \\
MOT17-04 & 73.3 & 85.5 & 87.8 & 47  & 3  & 755     & 2680    & 88.9    & 96.6    & 80   & 368  \\
MOT17-05 & 48.3 & 72.3 & 59.3 & 37  & 9  & 147     & 741     & 78.0    & 94.7    & 41   & 98   \\
MOT17-09 & 61.7 & 78.4 & 71.8 & 17  & 1  & 25      & 572     & 80.1    & 98.9    & 26   & 40   \\
MOT17-10 & 57.7 & 70.8 & 74.4 & 15  & 2  & 188     & 1480    & 75.0    & 95.9    & 59   & 240  \\
MOT17-11 & 62.9 & 67.1 & 72.6 & 21  & 9  & 492     & 984     & 78.2    & 87.8    & 12   & 70   \\
MOT17-13 & 58.3 & 68.2 & 75.9 & 29  & 1  & 442     & 540     & 82.9    & 85.5    & 22   & 108  \\
\midrule
Overall    & 63.5 & 73.6 & 76.4 & 178 & 37 & 2462    & 11418   & 78.8    & 94.5    & 331  & 1178 \\
\bottomrule
\end{tabular}
\caption{Detailed results on MOT17 validation split.}
\label{tab:metrics-on-MOT17-val-extended}
\end{table*}

\begin{table*}[]
\centering \footnotesize
\begin{tabular}{llllllllllll}
\toprule
Sequence & HOTA$\uparrow$ & MOTA$\uparrow$ & IDF1$\uparrow$ & MT$\uparrow$  & ML$\downarrow$  & FP$\downarrow$    & FN$\downarrow$     & Rcll$\uparrow$ & Prcn$\uparrow$ & ID Sw.$\downarrow$  & Frag$\downarrow$   \\ \midrule
MOT17-01 & 50.3 & 53.9 & 62.0 & 8   & 8   & 329   & 2602   & 59.7 & 92.1 & 42     & 118   \\
MOT17-03 & 67.8 & 88.9 & 83.3 & 129 & 1   & 3420  & 7978   & 92.4 & 96.6 & 218    & 1354  \\
MOT17-06 & 47.6 & 61.7 & 58.7 & 79  & 60  & 734   & 3668   & 68.9 & 91.7 & 116    & 352   \\
MOT17-07 & 47.1 & 64.7 & 56.8 & 22  & 11  & 618   & 5214   & 69.1 & 95.0 & 123    & 419   \\
MOT17-08 & 39.1 & 48.3 & 42.9 & 21  & 17  & 288   & 10439  & 50.6 & 97.4 & 195    & 491   \\
MOT17-12 & 55.7 & 60.0 & 66.5 & 37  & 21  & 636   & 2781   & 67.9 & 90.2 & 52     & 198   \\
MOT17-14 & 40.0 & 46.0 & 53.0 & 19  & 48  & 809   & 9042   & 51.1 & 92.1 & 127    & 653   \\ \midrule
Overall    & 58.9 & 73.7 & 71.8 & 945 & 498 & 20502 & 125172 & 77.8 & 95.5 & 2619   & 10755 \\ \bottomrule
\end{tabular}
\caption{Detailed results on MOT17 test split.}
\label{tab:metrics-on-MOT17-test-extended}
\end{table*}

\begin{table*}[]
\centering \footnotesize
\begin{tabular}{lllllllllllll}
\toprule
           & TETA$\uparrow$ & HOTA$\uparrow$ & MOTA$\uparrow$ & MOTP$\uparrow$ & IDF1$\uparrow$ & FP$\downarrow$    & FN$\downarrow$     & IDSw$\downarrow$ & MT$\uparrow$   & PT$\downarrow$   & ML$\downarrow$   & FM$\downarrow$    \\
\midrule
pedestrian & 58.4 & 46.3 & 55.6 & - & 56.0 & - & - & - & - & - & - & - \\
rider      & 52.8 & 44.1 & 43.8 & - & 59.8 & - & - & - & - & - & - & - \\
car        & 74.4 & 64.0 & 72.3 & - & 72.4 & - & - & - & - & - & - & - \\
truck      & 62.3 & 51.3 & 44.3 & - & 59.3 & - & - & - & - & - & - & - \\
bus        & 66.9 & 58.9 & 50.6 & - & 67.2 & - & - & - & - & - & - & - \\
train      & 23.3 & 1.8  & 0.0  & - & 2.5  & - & - & - & - & - & - & - \\
motorcycle & 52.1 & 46.4 & 34.7 & - & 58.1 & - & - & - & - & - & - & - \\
bicycle    & 52.8 & 41.1 & 32.7 & - & 47.9 & - & - & - & - & - & - & - \\
\midrule
Average    & 55.4 & 44.2 & 41.8 & 83.4 & 52.9 & 24580 & 113632 & 6360 & 8935 & 5862 & 3248 & 12707 \\
Overall    & 71.5 & 60.8 & 67.4 & 86.1 & 69.2 & 24580 & 113632 & 6360 & 8935 & 5862 & 3248 & 12707 \\
\bottomrule
\end{tabular}
\caption{Detailed results with a Swin-L backbone on BDD100K validation split.}
\label{tab:metrics-on-BDD100K-val-extendend}
\end{table*}

\begin{table*}[]
\centering \footnotesize
\begin{tabular}{lllllllllllll}
\toprule
           & TETA$\uparrow$ & HOTA$\uparrow$ & MOTA$\uparrow$ & MOTP$\uparrow$ & IDF1$\uparrow$ & FP$\downarrow$    & FN$\downarrow$     & IDSw$\downarrow$ & MT$\uparrow$   & PT$\downarrow$   & ML$\downarrow$   & FM$\downarrow$    \\
\midrule
pedestrian & 58.5 & 46.7 & 55.4 & - & 58.9 & - & - & - & - & - & - & - \\
rider      & 54.6 & 46.2 & 45.3 & - & 61.3 & - & - & - & - & - & - & - \\
car        & 74.6 & 64.7 & 73.4 & - & 73.8 & - & - & - & - & - & - & - \\
truck      & 60.1 & 49.0 & 39.8 & - & 58.1 & - & - & - & - & - & - & - \\
bus        & 63.4 & 54.6 & 44.4 & - & 61.4 & - & - & - & - & - & - & - \\
train      & 29.2 & 21.7 & 7.2 & - & 26.6 & - & - & - & - & - & - & - \\
motorcycle & 52.2 & 42.8 & 38.4 & - & 56.0 & - & - & - & - & - & - & - \\
bicycle    & 52.9 & 43.0 & 38.7 & - & 56.1 & - & - & - & - & - & - & - \\
\midrule
Average    & 55.7 & 46.1 & 42.8 & 80.6 & 56.5 & 48894 & 204063 & 10793 & 16917 & 10261 & 4953 & 22975 \\
Overall    & 71.4 & 61.1 & 67.7 & 85.7 & 70.5 & 48894 & 204063 & 10793 & 16917 & 10261 & 4953 & 22975 \\
\bottomrule
\end{tabular}
\caption{Detailed results with a Swin-L backbone on BDD100K test split.}
\label{tab:metrics-on-BDD100K-test-extendend}
\end{table*}